\documentclass[letterpaper]{article} % DO NOT CHANGE THIS
\usepackage{aaai25}  % DO NOT CHANGE THIS
\usepackage{times}  % DO NOT CHANGE THIS
\usepackage{helvet}  % DO NOT CHANGE THIS
\usepackage{courier}  % DO NOT CHANGE THIS
\usepackage[hyphens]{url}  % DO NOT CHANGE THIS
\usepackage{graphicx} % DO NOT CHANGE THIS
\usepackage{natbib}  % DO NOT CHANGE THIS AND DO NOT ADD ANY OPTIONS TO IT
\usepackage{caption} % DO NOT CHANGE THIS AND DO NOT ADD ANY OPTIONS TO IT
\usepackage{algorithm}
\usepackage{algorithmic}
\usepackage{aaai25}  % DO NOT CHANGE THIS
\usepackage{times}  % DO NOT CHANGE THIS
\usepackage{helvet}  % DO NOT CHANGE THIS
\usepackage{courier}  % DO NOT CHANGE THIS
\usepackage[hyphens]{url}  % DO NOT CHANGE THIS
\usepackage{graphicx} % DO NOT CHANGE THIS
\usepackage{natbib}  % DO NOT CHANGE THIS AND DO NOT ADD ANY OPTIONS TO IT
\usepackage{caption} % DO NOT CHANGE THIS AND DO NOT ADD ANY OPTIONS TO IT
\usepackage{algorithm}
\usepackage{algorithmic}
\usepackage{booktabs} 
\usepackage{newfloat}
\usepackage{listings}
\usepackage{soul}
\usepackage{color}

\usepackage{amsmath} 
\usepackage{multirow}
\usepackage{newfloat}
\usepackage{bibentry}
\usepackage{listings}
\DeclareCaptionStyle{ruled}{labelfont=normalfont,labelsep=colon,strut=off} % DO NOT CHANGE THIS
\floatstyle{ruled}
\newfloat{listing}{tb}{lst}{}
\floatname{listing}{Listing}
\title{Probability-Density-Aware Semi-supervised Learning}
\author{
  Shuyang Liu\thanks{The authors contribute equally to this paper.}^{,1}, \textbf{Ruiqiu Zheng}^{\star,1}, \textbf{Zhou Yu}\thanks{The corresponding author of this paper.}^{,1,3},\textbf{Yunhang Shen}^{\dagger,2}, \textbf{Zhou Yu}\thanks{The corresponding author of this paper.}^{,1,3},\\ 
  \textbf{Ke Li}^{2}, \textbf{Xing Sun}^{2},  \textbf{Shaohui Lin}^{1,3}\\
}

\affiliations{
    \textsuperscript{\rm 1} East China Normal University\\
  \textsuperscript{\rm 2} Youtu Lab, Tencent, Shanghai, China\\
  \textsuperscript{\rm 3} Laboratory of Advanced Theory and Application in Statistics and Data Science - MOE, China  \\
    zyu\text{@}stat.ecnu.edu.cn, shenyunhang01\text{@}gmail.com  }

\begin{document}

\section{Probability-Density-Aware Semi-supervised Learning}
\par
~
\par 

~~~\textbf{Shuyang Liu}, \textbf{Ruiqiu Zheng}\footnote{Equally contribution with the first author. }, \textbf{Yunhang Shen}\footnote{The corresponding author, Tencent Youtu Laboratory.}, \par ~~~~~~ \textbf{Zhou Yu}\footnote{The corresponding author, East China Normal University.},
  \textbf{Ke Li}, \textbf{Xing Sun}, \textbf{Shaohui Lin}
  \par
  ~~~~\textit{zyu\text{@}stat.ecnu.edu.cn,~~ shenyunhang01\text{@}gmail.com}
%\maketitle
\par
~
\par
\begin{abstract}
Semi-supervised learning~(SSL) assumes that neighbor points lie in the same category (\textit{neighbor assumption}), and points in different clusters belong to various categories~(\textit{cluster assumption}).
Existing methods usually rely on similarity measures to retrieve the similar neighbor points, ignoring \textit{cluster assumption}, which may not utilize unlabeled information sufficiently and effectively.
This paper first provides a systematical investigation into the significant role of probability density in SSL and lays a solid theoretical foundation for \textit{cluster assumption}.
To this end, we introduce a Probability-Density-Aware Measure~(PM) to discern the similarity between neighbor points.
To further improve Label Propagation, we also design a Probability-Density-Aware Measure Label Propagation~(PMLP) algorithm to fully consider the \textit{cluster assumption} in label propagation.
Last, but not least, we prove that traditional pseudo-labeling could be viewed as a particular case of PMLP, which provides a comprehensive theoretical understanding of PMLP's superior performance.
Extensive experiments demonstrate that PMLP achieves outstanding performance compared with other recent methods.

\end{abstract}

\section{Introduction}

Machine Learning's remarkable breakthroughs rely on constructing high-quality, complex labeled datasets.
However, dataset annotations are increasingly costly or even infeasible in many professional areas (\textit{e.g.}, medical and astronomical fields~\cite{chen2020self, xu2022one}). 
Semi-supervised learning~(SSL) has been proposed to mitigate the demand for large-scale labels by extracting information from unlabeled data with guidance from a few labeled data. 
The existing SSL methods are designed based on the widely accepted \textit{assumption of consistency},
which assumes that close data points probably have the same label and data points lie in the same cluster tend to have the same labels~\cite{zhou2003learning, chapelle2005semi, iscen2019label, li2020density, zhao2022lassl}. 
The latter prior that points from the same category should lie in the same cluster is specifically called \textit{cluster assumption}~\cite{zhou2003learning,iscen2019label, li2020density}. 

In SSL, the \textit{assumption of consistency} is usually considered by assigning the similarity between neighbor points.
An enormous similarity indicates that the points tend to be in the same category. 
Some works use similarity to train a consistent encoder, considering that neighbor input should have neighbor features~\cite{sohn2020fixmatch}. 
%The unsupervised consistent encoder can be followed by supervised fine-tuning~\cite{hinton2006fast, bengio2006greedy}. 
Contrastive learning learns consistent visual representations with unlabeled images~\cite{chen2020big}. 
Adversarial training~\cite{miyato2018virtual} and data augmentation~\cite{berthelot2019remixmatch, sohn2020fixmatch} are also widely applied to gain more training samples and improve the learning of representation. 
%The similarity is also used in pseudo-labeling.
High-confidence pseudo-labels are propagated with close neighbor points' predictions and labels~\cite{iscen2019label,zhao2022lassl,elezi2018transductive}. 
%KNN chooses better neighbors~\cite{nassar2023protocon}.
Combining consistent encoder and pseudo-labeling significantly improves the model performance~\cite{iscen2019label}.
Recently, the widely-used SSL framework often consists of supervised loss to take full consideration of labeled data, unsupervised loss to improve the quality of pseudo labels, and consistency loss to encourage a consistent encoder~\cite{zhao2022lassl,berthelot2019remixmatch,zheng2022simmatch}.
\begin{figure*}[t!]  
  
  \centering  
  \begin{minipage}{.92\textwidth}   
    \centering  
    \includegraphics[width=\linewidth]{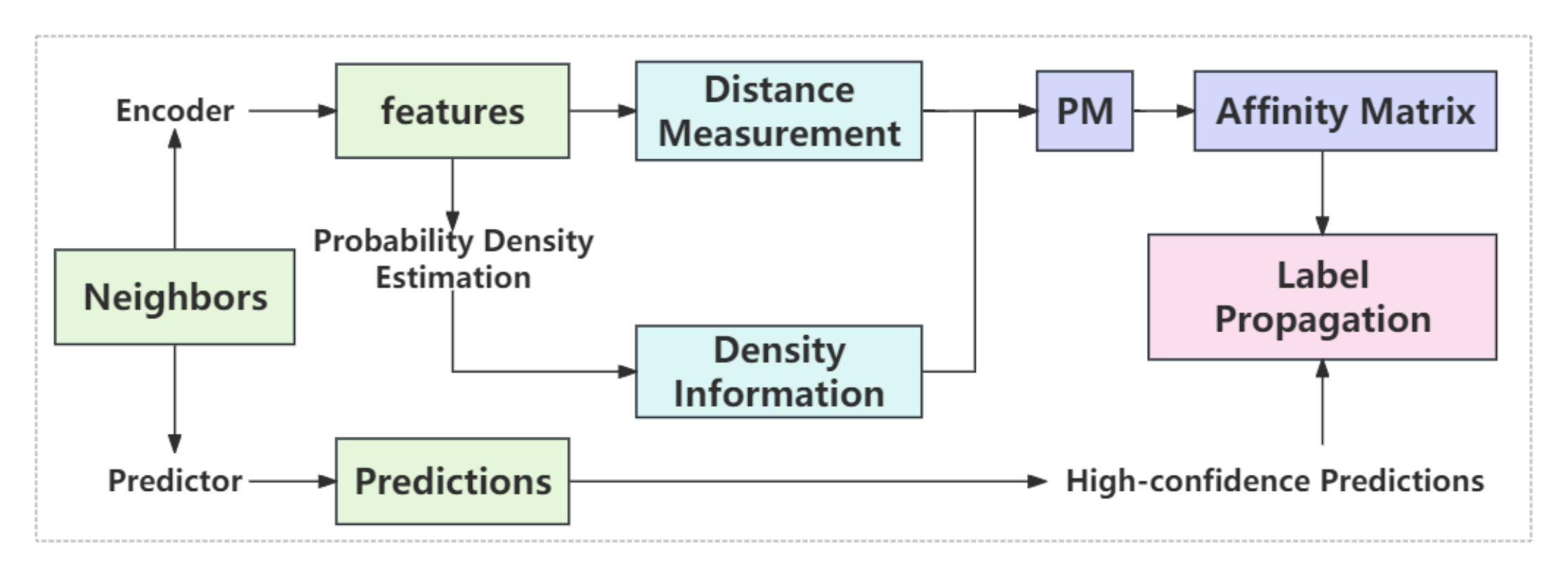}  
  \end{minipage}\hspace{0mm}     
    \caption{The procedure of PMLP.
    First, we select the neighbor points and extract their features.
    Then, we calculate the densities on the path; the density information is used to construct the Probability-density-aware Measure(PM). PM can fully consider the \textit{cluster assumption}.
    Finally, high-confidence predictions are used for pseudo-labeling with an affinity matrix.
    }\label{Fig1}
\end{figure*}

However, these works fail to fully consider \textit{assumption of consistency}, for they do not use the prior of \textit{cluster assumption}.
When calculating the similarity between points, their measurements seldom consider the cluster prior.
Considering that the points close to the decision boundary may be closer to points from other categories than some points in the same categories, these algorithms may be misled by wrongly assigning neighbors' labels. 
Our proposed solution, the Probability-Density-Aware Measurement(PM), has the potential to fully utilize the \textit{cluster assumption} through the incorporation of density information and could address the limitations of existing algorithms. 
A natural thought goes that for two points with different clusters, their path tends to traverse low-density regions. 
We statistically prove the thought and demonstrate that the path probably goes through low-density areas. The theorem indicates that density information significantly helps to consider the \textit{cluster assumption} and severs as the keystone of PM. 

With PM, we can better measure the similarity between points. We deploy PM into Label Propagation to select better neighbors and generate higher-confidence pseudo-labels. The new algorithm is named Probability-Density-Aware Label-Propagation(PMLP). 
To show PM's superiority and generality, we prove that traditional pseudo-labeling is a particular case of PMLP, which indicates that PMLP should always be better with just minor optimization. 
PM can be easily implemented to complement many popular SSL methods involving label propagation~\cite{zhou2003learning,iscen2019label,zhao2022lassl,elezi2018transductive}; we deploy PM to the best performance LPA algorithm ~\cite{zhao2022lassl} to implement the experiment.
Through comprehensive experiments on benchmark datasets, we show that PMLP achieves noticeable performance improvement compared to some of the most powerful approaches. 
%Specifically, PMLP surpasses the foremost competitor with a $1.11\%$ improvement and outperforms our baseline LASSL with a $3.52\%$ improvement on the CIFAR100 dataset with 400 labeled examples. 
%
In general, our contribution can be summarized into three points:
\begin{itemize}    
\item
We first propose a theoretical description to \textit{cluster assumption}, which reveals that probability density can help make full use of SSL's prior;
\item
We introduce a new Probablity-Density-Aware similarity measurement(PM) to fully consider SSL's cluster prior. Probability-Density-Aware Label-Propagation(PMLP) is proposed based on PM. We mathematically prove PMLP's superiority and generality;
\item
Extensive experiments are conducted to validate the effectiveness of PMLP. For example, on CIFAR100 with 400 labeled samples, we can surpass the baseline model by $3.52\%$ and surpass the second-best algorithm by $1.21\%$. 
\end{itemize}

%%%%%%%%%%%%%%%%%%%%%%%%%%%%%%%%%%%%%%%%%%%%%%%%%%%%%%%%%%%%%%%%%%%%%%%%%%%%%%

\section{Related Works}

\subsection{Consistency Regularization}

The \textit{assumption of consistency} is widely accepted in SSL. Its foundation is generating well-consistency models that exhibit similar features for similar inputs.
The works in~\cite{hinton2006fast,bengio2006greedy} improved consistency via self-training and finetuning the model with labeled data.
$\Pi-model$~\cite{laine2016temporal} reduces the consistency loss between images and their augmentations, or between outputs and temporal average outputs.
Mean-teacher~\cite{tarvainen2017mean} reduces consistency loss between current weights and temporal average weights.
Data augmentation can help to extract more feature information~\cite{miyato2018virtual,elezi2018transductive}.
%
%Data augmentation is expected to output similar features from different tiny-perturbed inputs and is widely used in SSL~\cite{miyato2018virtual,elezi2018transductive}.
%
Recent algorithm~\cite{chen2020self} involved contrastive learning to improve the model consistency, and LASSL~\cite{zhao2022lassl} generated class-aware contrastive learning by limiting the comparisons within the same class. 
However, these works fail to produce pseudo-labels directly.

\subsection{Pseudo-labeling}
Generating high-confidence pseudo-labels for unlabeled data is a widely used policy in SSL.
A simple yet intuitive approach considers high-confidence probability(higher than a threshold $\tau$) vectors to be accurate and assigns them pseudo-labels~\cite{sohn2020fixmatch}.
Freematch and Adsh~\cite{wang2022freematch,guo2022class} adaptive adjust threshold $\tau$ to improve the pseudo-label quality. 
Some early pseudo-labeling algorithms~\cite{lee2013pseudo, sajjadi2016regularization, shi2018transductive} rely heavily on predictions, while recent frameworks prefer graph-based label propagation yielding better consideration of \textit{assumption of consistency}~\cite{zhou2003learning,iscen2019label, zhao2022lassl, douze2018low}. 
Graph transduction game~\cite{elezi2018transductive,erdem2012graph} fixes the graph and lacks a weighting mechanism. 
%Learning by association~\cite{haeusser2017learning} uses two-step propagation on a graph between labeled and unlabeled examples.
%
Some statistical methods~\cite {zhou2003learning} take a reasonable consideration of \textit{assumption of consistency} by calculating the affinities via the first-order similarity between features and iteratively spread labels to their neighbors.
The strategy can extend to the high-dimensional feature space in deep learning~\cite{iscen2019label,zhao2022lassl}.
Comatch~\textit{et al.}~\cite{li2021comatch} introduces graphic information in contrastive learning and smooths pseudo-labels with a memory bank.
%
%Protocon~\textit{et al.}~\cite{nassar2023protocon} improves clustering through Label Refinement in K-Nearest Neighbors. 
However, these works fail to fully consider the \textit{cluster assumption}.

\subsection{Density-aware Methods}

Although the above techniques effectively select close neighbors, the \textit{cluster assumption} is not sufficiently considered, and they fail to distinguish neighbors' cluster affiliation directly. 
Some works intuitively point out that the decision boundary tends to be placed in a low-density region~\cite{chapelle2005semi,li2020density}, and this prior helps consider \textit{cluster assumption}. 
Wasserman \textit{et al.}~\cite{azizyan2013density} proposes a density-sensitive distance metric that provides a lower value when the points are located in different clusters.
Connectivity kernel~\cite{chapelle2005semi, bousquet2003measure} can utilize the detection of density variations along a path to determine whether two points lie in the same cluster.
These statistical methods can thoroughly consider \textit{cluster assumption}, but their optimization in high-dimensional space needs expensive computation.
Recently, DNA and DPLP~\cite{li2020density} prefer higher-density neighbors and construct a density-ascending path for pseudo-labeling. However, their density is cosine similarity, which harms the theoretical guarantee and statistical interpretability. 
Simmatch~\cite{zheng2022simmatch} tries to optimize pseudo labels with marginal distribution, but labeled data in SSL is few. 

Our algorithm aims at providing a general framework to fully consider the \textit{cluster assumption} by density-aware distance, and thus produce high-quality pseudo-labels. 
%
%Taherkhani~\textit{et al.}~\cite{Taherkhani2023wasersein} uses Wasserstein distance to match unlabeled clusters to the labeled clusters but falls short in concisely utilizing the geometric characteristics inherent in the \textit{cluster assumption}. 

\section{Methedology}

\subsection{A Statistical Explanation To \textit{Cluster Assumption}}

It has been widely accepted that with \textit{cluster assumption}, points in the same cluster tend to lie in the same category.  
In the previous work, however, it still lacks mathematical investigation.
This section presents an innovative statistical explanation demonstrating probability density is crucial in leveraging the \textit{cluster assumption}. 
To lay a theoretical analysis, without loss of generality, we assume the points in the feature space are widely accepted sub-gaussian random vectors. Then we have the \textbf{Theorem1}:

\textbf{Theorem1}
\textit{
Suppose two sub-gaussian random vectors $X_1, X_2$ with mean $\mu_1, \mu_2$.
Denote $x_1, x_2$ randomly sampled from $X_1, X_2$, and $\gamma(x_1, x_2)$~\cite{azizyan2013density} is a continuous finite path from $x_1$ to $x_2$.
For any $\tau \in R$, denote $p(x)$ is the probability density of point $x$ estimated by Kernel Density Estimator~(KDE), define event:}
\begin{equation*}
   C_\tau:\{\exists x \in \gamma(x_1,x_2), p(x)\leq \tau\}. 
\end{equation*}
 Then with $||\mu_1-\mu_2||_2^2 \rightarrow \infty$, we have: 
\begin{equation*}
    P(C_\tau) \rightarrow 1.
\end{equation*}

We can further prove that for two sufficiently dispersed $X_1,X_2$, almost every point on the straight line between $x_1,x_2$ traverses a low-density region.  

\textbf{Corollary1}
\textit{Specifically take $\gamma(x_1,x_2)$ in \textbf{Theorem1} as a straight line and $x\in\gamma(x_1,x_2)$ distributes uniformly. For $\forall \tau \in R$, with sufficient large $||\mu_1-\mu_2||_2^2$, we have:} 
\begin{equation*}
    \lim_{||\mu_1-\mu_2||_2^2 \rightarrow \infty}P(\{x:p(x) \leq \tau\ , x \in \gamma(x_1,x_2)\}) \rightarrow 1. 
\end{equation*}

We have statistically proved that density information helps consider the \textit{cluster assumption} besides traditional distance measures. Specifically, their path probably traverses a low-density region when the points lie in different clusters. 
We simply choose $\gamma(x_1, x_2)$ as the connecting line $l(x_1, x_2)$ to save computation and equidistant select $K$ samples $\{x_{i,j}^{1}, \dots ,x_{i,j}^{k}\}$ along $l(x_1, x_2)$. 
We estimate their densities $\{p(x_{i,j}^{1}),...,p(x_{i,j}^{k})\}$ with Kernel Density Estimator(KDE) and the generated density information is denoted as $\{I(p_{i,j})| \{p(x_{i,j}^{1}),...,p(x_{i,j}^{k})\}$. 
We hope $I(p_{i,j})$ tends large when $l(x_{i},x_j)$ traverses inside one cluster and tends small when $l(x_{i},x_j)$ traverses among different clusters.

With any traditional distance measure $D(x_i, x_j)$ (such as first-order similarity, Euclidean distance, and cosine similarity), we propose our density-aware distance measure:
\begin{equation*}  
\left\{  
\begin{alignedat}{2}  
&d(x_i,x_j) = I(p_{i,j})D^{-1}(x_i,x_j),~~~~ \mathrm{if}~~ i \neq j;\\ 
&d(x_i,x_j) = 0,~~~~~~~~~~~~~~~~~~~~~~~~~~~~~~~~~\mathrm{if}~~ i=j.\\ 
\end{alignedat}  
\right.
\end{equation*} 
We name $d(x_i,x_j)$ as Probability-Density-Aware Measure(PM). The \textbf{Theorem2} presents four choices of $I(p_{i,j})$ and proves PM's superiority. In PM, the distance measure $D(x_i,x_j)$ reflects the \textit{neighbor assumption} that neighbor points tend to have the same label, and the density information $I(p_{i,j})$ reflects the \textit{cluster assumption} that features lie in different clusters tend to have different labels, thus takes full consideration of \textit{assumption of consistency}.

\subsection{Probability-Density-Aware Measure Label-Propagation}
Traditional Label Propagation Algorithm~(LPA)~\cite{zhou2003learning,iscen2019label,zhao2022lassl} search neighbor points by a distance measure $D(x_i,x_j)$ including first-order similarity, cosine similarity, and Euclidean distance, which lack consideration of \textit{cluster assumption}. 
We propose Probability-Density-Aware Measure Label-Propagation~(PMLP) by introducing the PM $d(x_i,x_j) = I(p_{i,j})D^{-1}(x_i,x_j)$ into pseudo-labeling. 
Theorem 2 lays a theoretical foundation for the good performance of PMLP.

Suppose we have observations $O$ and labeled data $(X^L,Y^L)$, unlabeled data and its prediction $(X^U,p(X^U))$. 
As assumed in LPA, we propagate high-confidence predictions to low-confidence predictions by similarity measures between neighbor points.

As a familiar setting in LPA, we use KNN to choose $K$ nearest neighbors in feature space $O$, and then we get new $O^{LP} = \{o_{1}^{LP},...,o_{K}^{LP}\}$, and the corresponding $Y^{LP} = \{y_{1}^{LP},...,y_{K}^{LP}\} \in \{p(X^U)\}\cup\{Y^L\}$. 
%Then we reweight $Y^{LP}$ by threshold  $\tau$ and get $Y^{LP}_{K}$: 
Then we reweight $Y^{LP}$ by threshold  $\tau$: 
\begin{equation*}
   y_{t}^{LP,\mathrm{high}} = I(\max(y_{t}^{LP}) \geq \tau)y_{t}^{LP},
\end{equation*}
\begin{equation}
\label{2}
   y_{t}^{LP,\mathrm{low}} = I(\max(y_{t}^{LP}) < \tau)y_{t}^{LP},
\end{equation}
where $Y^{LP,\mathrm{high}}$ includes high-confidence predictions $y_{t}^{LP,\mathrm{high}}$ and ground-truth labels $Y^L$.
Denote any distance measurements between features as $D(o_{i}^{LP},{o_{j}^{LP}})$ in traditional LPA. 
To fully consider \textit{cluster assumption}, we introduce PM $d(x_i,x_j)$ to introduce the density information between two features $o_{i}^{LP}, {o_{j}^{LP}}$. 
% Then, we can get the new adjacent matrix with PM as the similarity measurement:
Then we get the reweighted adjacent matrix $W'$:
\begin{equation}  
\label{4}
W'_{i,j} = \left\{  
\begin{alignedat}{2}  
&0,                    & \quad &\text{if } i=j, \\  
&I(p_{i,j})D^{-1}(o_{i}^{LP},{o_{j}^{LP}}),                    & \quad &\text{if } i \neq j,  \\  
\end{alignedat}  
\right.
\end{equation}

Then, the pseudo-labels are generated by iteratively propagating the high-confidence predictions to their close and same-cluster neighbors based on the affinity matrix $W'$.
Denote $\widehat Y^{'LP}(i)$ implies the propagated pseudo-label in $i$-th iteration.
Note $D$ the diagonal matrix with $(i, i)$-th equal to the sum of the $i$-th row of $W'$, and iteratively propagate the label information to unlabeled points:
\begin{equation}
\label{5}
   \widehat Y^{'LP}(i) = \alpha D^{-\frac{1}{2}}W'D^{-\frac{1}{2}}\widehat Y^{LP}(i-1) + (1-\alpha)Y^{LP,\mathrm{high}}.
\end{equation} 
We get the final optimal form $\widehat Y^{'LP}$:
\begin{equation*}
   \widehat Y^{'LP} = (I-\alpha D'^{-\frac{1}{2}}W'D'^{-\frac{1}{2}})^{-1}Y^{LP,\mathrm{high}}.
\end{equation*}
And the final pseudo-label $\widehat Y^{*LP}$ is combined by $\widehat Y^{'LP}$ and our prediction $Y^{LP,low}$:
\begin{equation}
\label{6}
   \widehat Y^{*LP} = \eta \widehat Y^{'LP} + (1 - \eta)(Y^{LP,\mathrm{low}}).
\end{equation}

The superiority and generalization of PMLP lie in its good theoretical guarantee: whatever the choice of $D(o_{i}^{LP},{o_{j}^{LP}})$ and loss function $L(Y^{P},\widehat Y^{*LP})$, 
we prove that, when $I(p_{i,j})$ is chosen as:
\begin{equation*}  
\label{equ7}
\left\{  
\begin{alignedat}{4}  
&I(p_{i,j}) = \max\{p_{t}(i,j),t=1,2,...,k\},                    & \quad &\text{if } i=j, \\ 
&I(p_{i,j}) = \min\{p_{t}(i,j),t=1,2,...,k\},                    & \quad &\text{if } i \neq j,  \\
&I(p_{i,j}) = \operatorname{avg}(\{p_{t}(i,j),t=1,2,...,k\}),                    & \quad &\text{if } i \neq j,  \\
&I(p_{i,j}) = Q_t(\{p_{t}(i,j),t=1,2,...,k\})),                    & \quad &\text{if } i \neq j,  \\
\end{alignedat}  
\right
.
\end{equation*} 
where $\operatorname{avg}(\cdot)$ denotes the mean and $Q_t(\cdot)$ denotes the $t-quantile$, in our implemention, we take it as the median. 
The procedure of PMLP is in pseudo code~\ref{pseudo_code}.

\begin{algorithm}[tb]
\caption{The algorithm for density-aware label propagation}
\label{pseudo_code}
\textbf{Input}:   %$K$ nearest neighbor features $O^{LP}$ for constructing $W$,
                  
                  %corresponding $K$ labels from the latest iter $Y^{LP}$,  
                  $K$, $Y^{LP}$, $O^{LP}$,
                  threshold $\tau$,
                  %$k$ for choosing neighbors for KDE, 
                  k, n,
                  %$\alpha$ for propagating pseudo-label $Y^{'LP}$, 
                  $\alpha$,
                  %$\eta$ for mixup high-quality prediction $Y^{LP}$
                  $\eta$.\\
\textbf{Output}: density-aware pseudo-label $\widehat Y^{\star LP}$\\
\begin{algorithmic}[1] %[1] enables line numbers
\STATE Choose $K$ nearest neighbor points in features $O^{LP}$;

\STATE Seperate $Y^{LP,\mathrm{high}}$ and $Y^{LP,\mathrm{low}}$ by Eq.~\ref{2} with $\tau$;   

\STATE  %For points in \( Y^{LP,\mathrm{low}} \) and \( Y^{LP,\mathrm{high}} \), calculate the distance by Eq.~\ref{3} to obtain \( W_{i,j} \).
For points in \( Y^{LP,\mathrm{low}} \) and \( Y^{LP,\mathrm{high}} \), calculate the distance $D(x_i,x_j)$.

\STATE Get k-equal division points on the connecting line, for each point in ${o_{i,j}^{1},...,o_{i,j}^{k}}$, select $n$ neighbors to calculate density $p_{i,j}$ with KDE;

\STATE Reweight $W$ with $I(p_{i,j})$ as Eq.~\ref{4}, get $W'$;

\STATE Propagating the label with $W'$, $\alpha$ by Eq.~\ref{5}, get $Y^{'LP}$;

\STATE Mix the propagated label $Y^{'LP}$ and $Y^{LP}$ by Eq.~\ref{6} with $\eta$, get final density-aware pseudo-label $\widehat Y^{*LP}$.
\end{algorithmic}
\end{algorithm}

We prove that PMLP can consistently outperform traditional LPA. Traditional LPA is a particular case of PMLP, indicating both PMLP's superiority and reasonableness. 

\textbf{Theorem2.}
\textit{
Denote the model parameter in one iteration as $\theta$.
%The encoder is $h(\cdot)$, and the predictor is $p(\cdot)$. 
And let $\mathcal{L}_{U}$ be the unsupervised loss between predictions and pseudo-labels.
%Denote $O^{LP}$ as the set of features and $Y^{P}$ as the prediction vector.
%The $i$-th element of $Y^{P}$ is then $y_{i}^{P} = f\cdot h(x_{i})$.
Denote $Y^{P}$ as the set of predictions, $\widehat Y^{*LP}$ and $\widehat Y^{LP}$ be the pseudo-labels generated by PMLP and traditional LPA with any 
 $D(\cdot,\cdot)$~\cite{zhou2003learning,iscen2019label,zhao2022lassl}. 
Suppose we calculate the probability density by KDE with exponential kernel and bandwidth $h$. 
Then $\mathcal{L}_{U}$ satisfies:}
\begin{equation*}
\min_{\theta \in \Theta, h\in R}\mathcal{L}_{U}(Y^{P}, \widehat Y^{*LP}) \leq \min_{\theta \in \Theta}\mathcal{L}_{U}(Y^{P}, \widehat Y^{LP}).
\end{equation*}

\begin{table*}[t]
  \centering
   \fontsize{8}{11}\selectfont
  \begin{tabular}{lllllllll}
    \toprule
    \hline
   \multirow{2}{*}{Method}&\multicolumn{2}{c}{CIFAR10}&\multicolumn{2}{c}{CIFAR100}&\multicolumn{2}{c}{SVHN}&\multicolumn{2}{c}{STL-10} \\
   
    \cmidrule(r){2-3}  \cmidrule(r){4-5}  \cmidrule(r){6-7} \cmidrule(r){8-9}
     & ~~40 labels   &~250 labels   &~ 400 labels   &2500 labels      &~~40 labels &~250 labels &~~40 labels &1000 labels\\
    \midrule
    Pseudo-label\hfill (2013) &25.39±0.26 &53.51±2.20  &12.55±0.85  &42.26±0.28            &25.39±5.6       &79.79±1.09            &25.31±0.99   &67.36±0.71        \\ %freematch
    Mean-Teacher\hfill (2017) &29.91±1.60 &62.54±3.30   &18.89±1.44    &54.83±1.06         &33.81±3.98      &96.43±0.11            &29.28±1.45   &66.10±1.37        \\ %freematch
    MixMatch\hfill (2019)     &63.81±6.48   &86.37±0.59   &32.41±0.66   &60.24±0.48        &69.40±3.39       &96.02±0.23            &45.07±0.96   &88.3±0.68          \\  %freematch
    UDA\hfill (2019)          &89.33±3.75    &94.84±0.06   &53.61±1.59   &72.27±0.21       &94.88±4.27      &98.01±0.02            &62.58±3.44   &93.36±0.17      \\ %softmatch epass
    ReMixMatch \hfill (2019)         &90.12±1.03 &93.70±0.05 &57.25±1.05 &73.97±0.35       &79.96±3.13      &97.08±0.48            &67.88±6.24   &93.26±0.14           \\ %freematch
    FixMatch \hfill (2020)           &92.53±0.28 &95.14±0.05 &53.58±0.82 &71.97±0.16       &96,19±1.18      &98.03±0.01            &64.03±4.14   &93,75±0.33          \\ %freematch epass
    ReFixMatch\hfill (2023)         &95.06±0.01  &95.17±0.05 &53.88±1.07  &72.72±0.22   &71.40±4.21       &94.26±0.3             &97.85±1.23   &98.11±0.03           \\%refixmatch
    Softmatch \hfill  (2023)       &94.89±0.14   &95.04±0.09    &62.40±024  &73.61±0.38     &97.54±0.24      &97.99±0.01            &77.78±3.82   &94.21±0.15\\
    EPASS \hfill (2023)           & 94.69±0.1   &94.92±0.05   & 61.12±0.24  &74.32±0.33    &97.02±0.02      &97.96±0.02            &84.39±2.48   &94.06±1.42             \\
    Freematch \hfill  (2023)       &  95.1±0.04  & 95.12±0.18   &62.02±0.42  &73.53±0.2    &\textbf{98.03±0.02}      &98.03±0.01   &84.46±0.55   &94.37±0.15          \\
    Shrinkmatch \hfill  (2023)    & ~~~~ 94.92    &~~~~95.26      &~~~~64.64    &~~~~74.83     &~~~~97.49        &~~~~98.04              &~~~~\textbf{85.98}     &~~~~94.18\\
    LASSL \hfill (2022,baseline)   &95.07± 0.78  &95.71 ±0.46   &62.33±2.69  &74.67± 0.65    &96.91±0.52      &97.85± 0.13         &81.57±0.36        &94.23±0.26          \\
    \midrule
    PMLP+LASSL \hfill (Ours)             & \textbf{95.42±0.32}   &\textbf{95.76±0.14}   & \textbf{65.85±0.8}  &\textbf{75.27±0.19}    &97.85±0.31 &\textbf{98.10±0.05} &85.53±1.92 &\textbf{94.53±0.53}\\
    Outperform than LASSL          &~~~  + 0.35  &~~~ + 0.05   &~~~+ \textbf{3.52}  &~~~+ \textbf{0.60}    &~~~+ \textbf{0.94}      &~~~+ 0.25   &~~~+ \textbf{3.96}   &~~~+0.30          \\
    \hline
    \bottomrule
   
  \end{tabular} \caption{
  Performance comparison on CIFAR10, CIFAR100, SVHN, and STL-10.
  We show the mean accuracy from $5$ experiments and the standard deviation.
  }\label{tabel1}
\end{table*}

\begin{figure*}[t!]  
  \centering  
  \begin{minipage}{.33\textwidth}  
    \centering  
    \includegraphics[width=\linewidth]{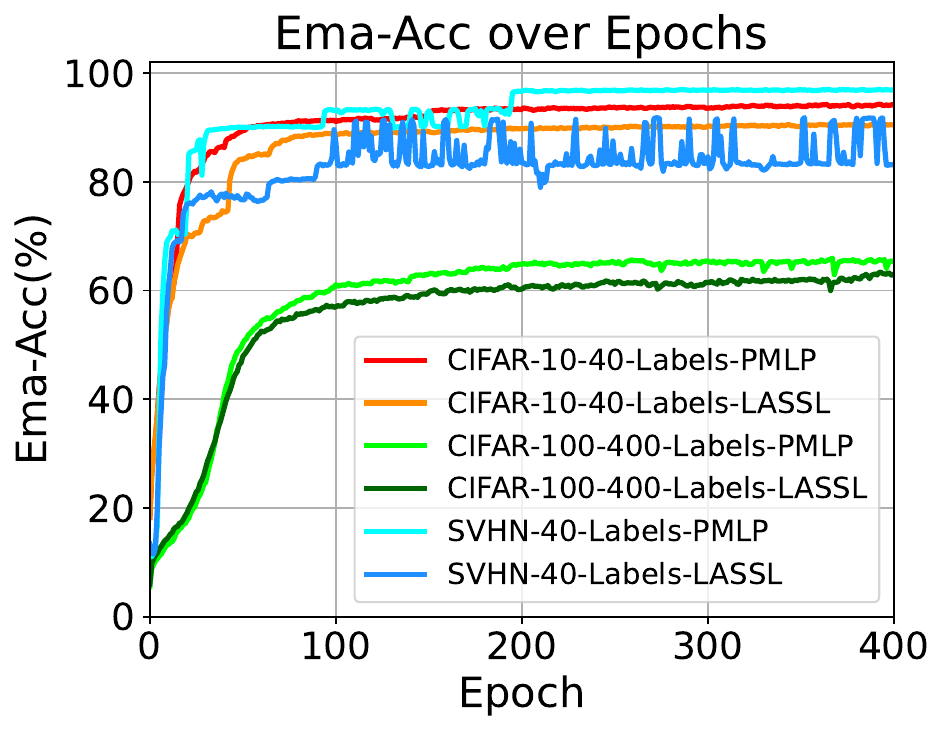}  
  \end{minipage}  
  \begin{minipage}{.33\textwidth}  
    \centering  
    \includegraphics[width=\linewidth]{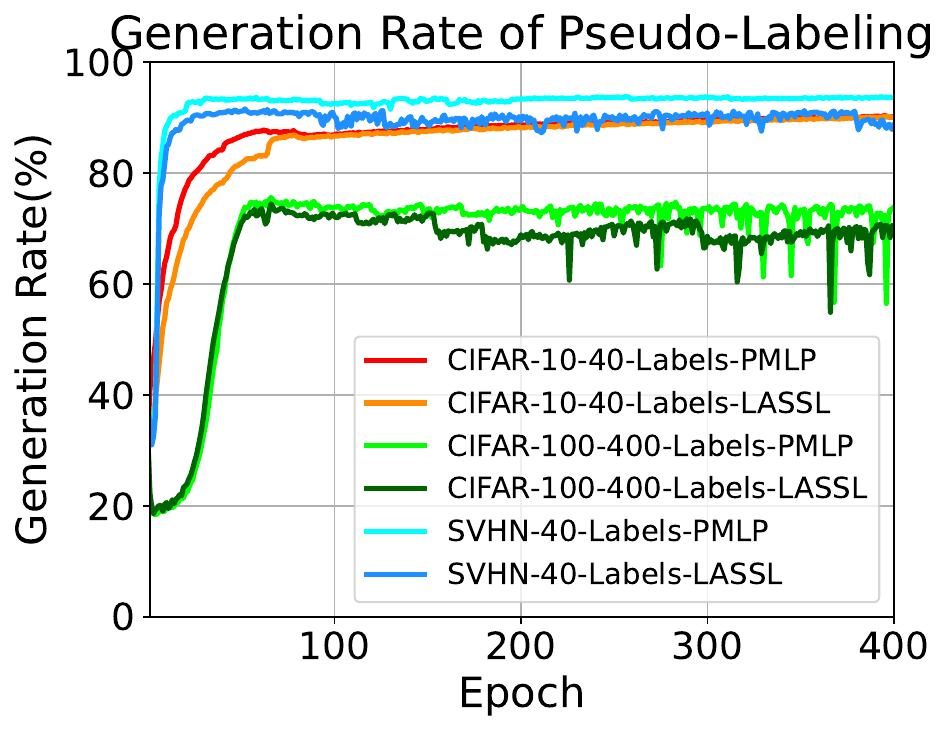}   
  \end{minipage} 
  \begin{minipage}{.33\textwidth}  
    \centering  
    \includegraphics[width=\linewidth]{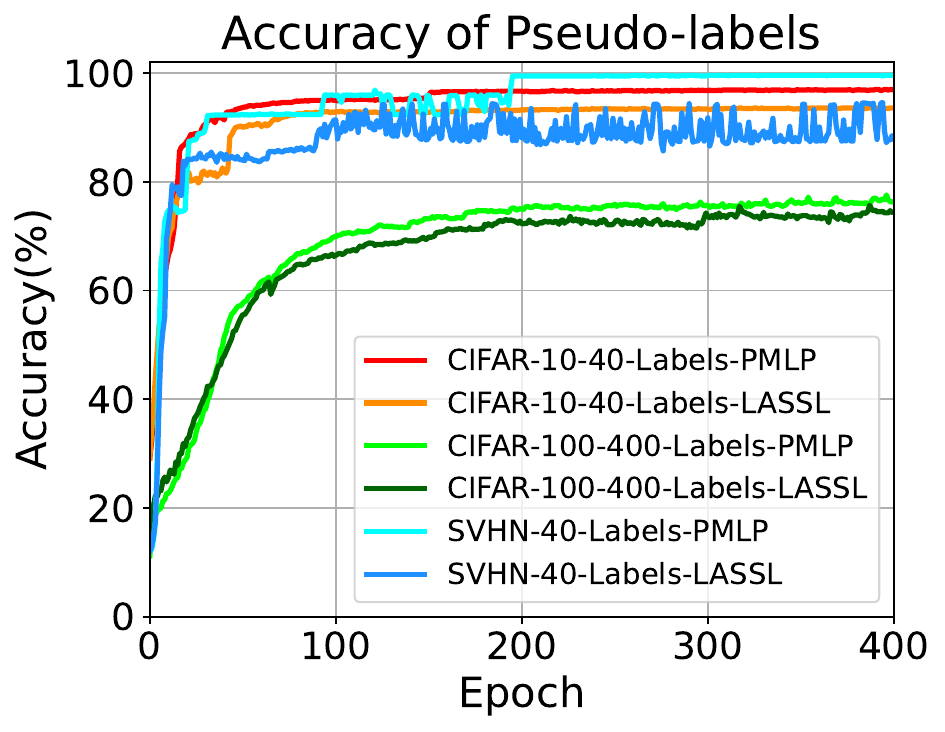} 
  \end{minipage}  
  \caption{Left: ema-accuracy of models. Middle: rate of high-quality pseudo-labels. Right: rate of correct high-quality labels.}  
  \label{Fig3}  
\end{figure*} 

\subsection{The Competitive Pseudo-Labeling}
%Pseudo-labeling relies heavily on excellent representations to detect the similarity between points.
Recent pseudo-labeling algorithms combine a consistent encoder, considering that a better encoder can generate 
better representations and facilitate the pseudo-labeling~\cite{iscen2019label,zhao2022lassl,berthelot2019remixmatch,sohn2020fixmatch,zheng2022simmatch}.
We inherit this framework in the implementation of PMLP. 
We choose the Label-guided Self-training approach to Semi-supervised Learning~(LASSL)~\cite{zhao2022lassl} as our baseline. 
As the best-performing label-propagation algorithm in recent years, LASSL still behaves dissatisfiedly compared with other recent algorithms. 
We hope deploying PMLP with the LASSL framework will allow it to surpass recent algorithms, thus demonstrating the full potential of label propagation under \textit{cluster assumption}.

In our implementation, we update the model with three losses: the supervised cross-entropy loss $\mathcal{L}_{S}$, the class-aware contrastive~(CACL) loss $\mathcal{L}_{C}$~\cite{zhao2022lassl}, and the unsupervised cross-entropy loss $\mathcal{L}_{U}$.

Denote $\theta$ as the model's parameters, labeled dataset $X^{L}=\{X_{L}, Y_{L}\}$, the model takes a consideration of labeled data by minimizing $\mathcal{L}_{S}$:
\begin{equation*}
    \mathcal{L}_{S}(X^{L}; Y^{L}; \theta) := CE(X^{L}; Y^{L}; \theta).
\end{equation*}
Denote the weak and strong augmentation of input $x$ as $a(x), A(x)$, denote $z^{L}_{i} = G(a(x_{i}))$ and $z^{U}_{j} = G(A(u_{j}))$ as the output features for both labeled and unlabeled data.
In the $t$-th iteration, note $Y_{t-1}=\{y_{1}^{L},...,y_{B}^{L}\} \cup \{y_{1}^{U},...,y_{\mu B}^{U}\}$ all predictions of labeled and unlabeled data,
for $y_{i},y_{j}\in Y_{t-1}$, introduce the instance relationships to help the contrastive learning:
\begin{equation*}  
w_{i,j} = \left\{  
\begin{alignedat}{2}  
&1,                    & \quad &\text{if } i=j, \\  
&0,                    & \quad &\text{if } i \neq j \text{ and } y_{i} \cdot y_{j} \leq \epsilon, \\  
&y_{i} \cdot y_{j},    & \quad &\text{if } i \neq j \text{ and } y_{i} \cdot y_{j} \geq \epsilon,
\end{alignedat}  
\right.
\end{equation*}  
where $ \epsilon $ is a hyper-parameter determining the confidence that two instances belong to the same category. 

Then we get $\mathcal{L}_{C}$ to help learning a good representation:
\begin{equation*}
    \mathcal{L}_{C} = -\Sigma_{i=1}^{|Y_{t-1}|}\log(\frac{\Sigma_{j=1}^{|Y_{t-1}|}w_{i,j}\exp(\frac{z_{i}\cdot z_{j}}{T})} {\Sigma_{j=1,j\neq i}^{|Y_{t-1}|}\exp(\frac{z_{i}\cdot z_{j}}{T})}),
\end{equation*}
where $T$ is temperature hyper-parameter~\cite{chen2020self}.

Denote $Y^{U}= \{y_{1}^{U}, \dots, y_{\mu B}^{U}\}$ the predictions of unlabeled data Unsupervised loss $\mathcal{L}_U$ encourages predictions to be consistent with PMLP's confidence pseudo-labels $Y^{*LP}$, which in turn encourages consistency encoder and helps to improve the pseudo-labeling: 
\begin{equation*}
    \mathcal{L}_{U} = I(y_{j}^{U} \geq \tau)H(Y^{U},\widehat Y^{*LP}).
\end{equation*}
Combine the three losses, we aim to update the model's parameter $\theta$ to minimize the loss function $L$:
\begin{equation*}
    \min_{\theta \in  \Theta} \mathcal{L} = \min_{\theta \in  \Theta}(\mathcal{L}_{S} + \mathcal{L}_{U} + \mathcal{L}_{C}).
\end{equation*}

\begin{figure*}[t!]
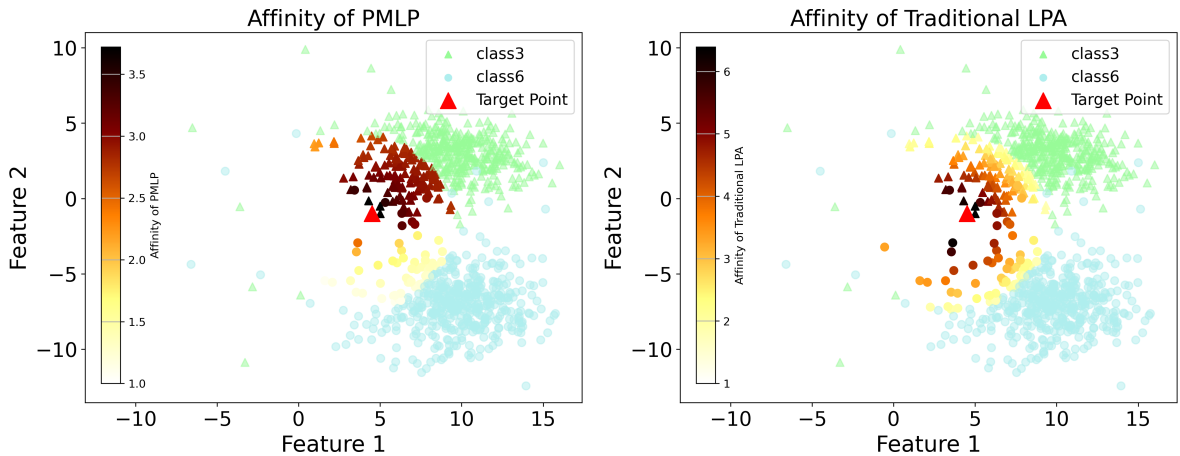

  \centering  

  \begin{minipage}{.44\textwidth}   
    \centering  
    \includegraphics[width=\linewidth]{PMLP_814.pdf}  
  \end{minipage}   
  \begin{minipage}{.44\textwidth}  
    \centering  
    \includegraphics[width=\linewidth]{LPA_814.pdf}  
  \end{minipage}    
  \caption{
  Different colors represent different distances between the target point and neighbor points.
  Black represents a close distance and a higher affinity.
  The left one chooses PM as the distance measure, and the right one chooses traditional first-order similarity.
  PMLP tends to choose neighbors within one cluster, and LPA equally chooses neighbors with different clusters.
  }  \label{Fig4}
\end{figure*}

\begin{table}
  \centering
 \fontsize{9}{10}\selectfont
  \begin{tabular}{llllll}
    \toprule \hline
    Epoch        & 0-4   &5-9   &10-14   &15-19    \\
    \midrule
    Time(s), PMLP  &748.4    &747.3   &747.1      &747.0        \\
    Time(s), KDE   &3733.0     &4244.0    &4166.6      &4189.0   \\
    \hline
    \bottomrule 
  \end{tabular}
  \caption{Running time for using KDE and our PMLP to compute the density. }\label{table2}
\end{table}

\begin{table*}
\centering
\fontsize{8.1}{12}\selectfont
\renewcommand{\arraystretch}{1.2}
\begin{tabular}{c|c|c|c|c|c|c}
\toprule
 \hline
 \multirow{2}{*}{Dataset}&\multirow{2}{*}{Method} & \multirow{2}{*}{Average Time \small(s/epoch)} &\multicolumn{4}{c}{Average Time Across Different Iterations \small(s/epoch)} \\
\cline{4-7}
 &  & & 1-256 & 257-512 & 513-768 & 769-1024 \\
\hline
\multirow{2}{*}{STL-10} & LASSL & 615.21 & 593.81 & 625.56 & 618.04 & 623.40 \\ 
\cline{2-7} 
 & LASSL+PMLP & 648.08(\small\textbf{+5.34\%}) & 734.59(\small\textbf{+23.71\%}) & 652.42(\small\textbf{+4.29\%}) & 606.64(\small\textbf{-1.85\%}) & 598.69(\small\textbf{-3.96\%}) \\ \cline{2-7} 
 %& Time Increase & \textbf{5.34\%} & \textbf{23.71\%} & \textbf{4.29\%} & \textbf{-1.85\%} & \textbf{-3.96\%} \\ 
 \hline
\multirow{2}{*}{CIFAR10} & LASSL & 211.41 & 255.52 & 196.10 & 196.66 & 197.35 \\ 
\cline{2-7} 
 & LASSL+PMLP & 217.94 (\small\textbf{+3.09\%}) & 260.05 (\small\textbf{+1.77\%}) & 203.72 (\small\textbf{+3.89\%}) & 203.52 (\small\textbf{+3.89\%}) & 204.46 (\small\textbf{+3.60\%}) \\ 
 \cline{1-7} 
 %& Time Increase & \textbf{3.09\%} & \textbf{1.77\%} & \textbf{3.89\%} & \textbf{3.49\%} & \textbf{3.60\%} \\ 
 % \hline
 \bottomrule
\end{tabular}
\caption{Overall average time and average time of different iterations for different methods on STL-10 and CIFAR-10 datasets.}
\label{table3}
\end{table*}

\section{Experiments}
\subsection{Experiment Setup}\footnote{Code: \url{https://github.com/sdagfgaf/Probability-Density-aware-Semi-supervised-Learning}}
%In this section, we describe the experimental setup and hyper-parameters for PMLP.
We present comprehensive experiments of PMLP across extensive datasets, including SVHN~\cite{krizhevsky2009learning}, CIFAR10~\cite{krizhevsky2009learning}, and CIFAR100~\cite{netzer2011reading}, STL-10~\cite{AdamCoates2011}.
Following the standard protocol of SSL~\cite{zhao2022lassl,zheng2022simmatch,wang2022freematch,chen2023softmatch,yang2023shrinking}, we randomly select $40$ and $250$ labeled samples from SVHN and CIFAR10. 
For CIFAR100, $400$ and $2,500$ labeled samples are randomly selected.
For STL-10, we select $40, 1000$ labeled samples. 
For SVHN and CIFAR10, we use WideResNet-28-2 as the encoder to generate representations.
For CIFAR100, encoder is WideResNet-28-8.
For STL-10, encoder WideResNet-37-2. 
The predictor is a one-linear layer network.
In the CACL, we calculate $z_{i}$ with a $2$-layer MLP projector. 
The batch size includes $64$ unlabeled and $448$ labeled points.
In the label-propagation process, we select $1.5 \times N$ nearest neighbors, where $N$ represents the dataset category. 
In PMLP, we use $\alpha = 0.8$, $\eta = 0.2$, and $\tau = 0.95$. 
In CACL, $\epsilon$ is set to $0.7$. Optimization is performed using the SGD optimizer with a momentum of $0.9$ and a weight decay of $5 \times 10^{-4}$. 
The learning rate follows a cosine decay schedule. 
For KDE, we chose a bandwidth of $h = 5$ for SVHN, CIFAR10, and CIFAR100, and $h = 3$ for STL-10. 
In KDE, we select $512$ points for CIFAR10 and SVHN, and the nearest $45$ points for CIFAR100 and STL-10.

\subsection{Results On SVHN, CIFAR10, CIFAR100, STL-10}
Tab.~\ref{tabel1} compares the average testing accuracy of PMLP over $5$ runs against both classical algorithms and recent SOTA SSL approaches~\cite{zheng2022simmatch,wang2022freematch,chen2023softmatch,yang2023shrinking,nguyen2024debiasing,zhao2022lassl,nguyen2023boosting}. Here we set $I(p_{i,j})= \operatorname{avg}(\{p_{t}(i,j),t=1,2,...,k\})$, $K=1$.
PMLP consistently outperforms other SOTA approaches on CIFAR10 and CIFAR100 under all settings. With 400 labeled samples in CIFAR100, PMLP achieves an accuracy of $65.85\%$, surpassing the second-best ShrinkMatch~\cite{yang2023shrinking} by $1.21\%$ and improving our baseline LASSL with $3.52\%$. We also improved our baseline LASSL with $0.6\%$ thus surpassing the second-best ShrinkMatch with $0.44\%$.
On CIFAR10, PMLP always surpasses ShrinkMatch by more than $0.5\%$. 
With $40$ labeled samples in STL-10, PMLP reaches an accuracy of $85.53\%$, improving our baseline model LASSL with $3.96\%$; 
%For CIFAR10, with $40$ labeled samples, PMLP achieves an accuracy of $95.42\%$, outperforming the second-best FreeMatch~\cite{wang2022freematch} by $0.32\%$. 
for SVHN with $40$ labels, our PMLP can outperform the baseline LASSL with $0.94\%$, while still competitive compared with the SOTA.

\subsection{PMLP Generates Better Pseudo-labels}

In this section, we design experiments to show that PMLP generates better pseudo-labels. 
%Given a threshold $\tau$, the rate of high-confidence pseudo-labels above $\tau$ reflects the number of confidence pseudo-labels. The accuracy of pseudo-labels also significantly reflects the training. 
Fig.~\ref{Fig3} shows that LASSL and PMLP generate almost the same number of high-confidence pseudo-labels, but PMLP's pseudo-labels are more accurate. It directly explains PMLP's superior performance while implying that traditional LPA brings more misleading wrong labels into training, harming the model's performance. 

We also visually show the PMLP's selected neighbors in the feature space. 
%To explain why PMLP can generate more correct labels, we visually show the selected neighbors in feature space. 
By \textit{Theorem1}, with the same distance, we tend to choose neighbors in the same cluster and reject the neighbors in the different clusters. 
We implement PMLP and classical LPA with $40$ CIFAR10 labeled samples and output $64$-dimensional features. 
Principal component analysis~(PCA) is applied, and we select the two most important principal components and plot them in Fig.~\ref{Fig4}. 
We select the target point from class $3$ and its nearest $200$ neighbors.
In Fig.~\ref{Fig4}, we deploy PMLP and the classical LPA~\cite{zhou2003learning,iscen2019label,zhao2022lassl} and color the neighbors according to the value of $d(x_i,x_j)$ and $D(x_i,x_j)$. 
The deeper color represents a closer distance and more significant affinity. 
It shows that PMLP tends to choose neighbors in the same cluster. 
In PMLP, we introduce density information $I(p_{i,j})$ into $d(x_i, x_j)$; some neighbors in the other clusters are chosen but assigned with a lower affinity and tend to be diminished in the LPA process. 
%Conversely, LPA only considers the distance between features and tends to choose neighbors uniformly from two clusters.

\subsection{Acceleration Results Of PMLP}
We introduce the density information $I(p_{i,j})$, and the probability density is calculated with KDE. 
A natural thought goes that KDE is expensive to compute. Considering we choose $N$ samples on the connecting line and pick up $n$ support points each time for KDE, $O(N^2 kn)$ computation is needed. 
The KDE from Sklearn runs only on the CPU, leading to slower calculations. 
We use the exponential kernel $K(\cdot)$ in KDE to promote the divergence and design a GPU-based KDE, which can reach $\times 5$ acceleration Tab.~\ref{table2}. The details can be seen in the Appendix\footnote{Appendix: \url{https://arxiv.org/submit/6072244/view}}. 
We further compare the average time consumption in training between PMLP and our baseline LASSL in Tab.~\ref{table3}, indicating that PMLP does not cost too much time(less than about 5\%) than traditional LPA.

\begin{table}  
\centering  
\fontsize{9.6}{14}\selectfont  
\begin{tabular}{c|c|c|c|c}  
\toprule  
\hline  
Bandwidth & {$I(p_{i,j})$} & \textit{K}=1 & \textit{K}=3 & \textit{K}=5 \\  
\hline  
\multirow{4}{*}{h=5} & $\min(p_{k}(i,j))$ & 94.75 & \textbf{88.82} & \textbf{93.70} \\   
& $\max(p_{k}(i,j))$ & 95.42 & 95.19 & \textbf{87.56} \\    
& $\operatorname{avg}(p_{k}(i,j))$ & 95.61 & 95.35 & 95.19 \\  
& $Q_t(p_{k}(i,j))$ & 94.97 & 95.62 & 95.45 \\  
\hline  
\multirow{1}{*}{$h\rightarrow +\infty$, LASSL} & None & \multicolumn{3}{c}{95.30} \\   
\hline  
\bottomrule  
\end{tabular}  
\caption{Comparison of different $K$ and $I(p_{i,j})$ on CIFAR10 dataset with 40 labeled data.}  
\label{table4}  
\end{table}

\begin{table}[t] 
  \centering
\fontsize{8.1}{12.5}\selectfont
  \begin{tabular}{c|c|c|c|c|c|c}
    \toprule
    \hline
    Bandwidth     & \textit{h}=0.01   &\textit{h}=0.5   &\textit{h}=5   &\textit{h}=100      & \textit{h}=$+\infty$ &LASSL  \\
    \midrule
    Acc           &63.34    &65.42   &\textbf{66.5}      &64.32   &62.50        &62.33\\
    Density Ratio &1.73     &1.39    &1.09      &1.009   &1            &1\\
    \hline
    \bottomrule 
  \end{tabular}
  \caption{Ablation study on CIFAR100 with $400$ labeled data.}
  \label{table5}
\end{table}

\begin{table}[t] 
  \centering
\fontsize{9.1}{12.5}\selectfont
  \begin{tabular}{c|c|c|c|c|c}
    \toprule
    \hline
    Neighbor points(\textit{N})     & \textit{N}=3   &\textit{N}=8   &\textit{N}=10   &\textit{N}=15      & \textit{N}=20  \\
    \midrule
    Acc           &82.71    &94.21   &95.04      &\textbf{95.55}   &95.21        \\
    \hline
    \bottomrule 
  \end{tabular}
  \caption{Ablation study with $K$ neighbors on CIFAR10.}
  \label{table6}
\end{table}

\subsection{Ablation Studies}
\label{sec4.4}

\subsubsection{The Effect Of $I(p_{i,j})$}
In Tab.~\ref{table5}, we compare PMLP's performance with different $I(p_{i,j})$ and different $K$ points on the connecting line. 
We deploy the experiment with the CIFAR10 dataset, 40 labeled samples. 
%Intuitively, with larger $K$, the accuracy tends to be stable and superior. 
We use simplify notations for $I(p_{i,j})$, such as $ \operatorname{avg}(\{p_{t}(i,j),t=1,2,...,k\})=\operatorname{avg}(p_{t}(i,j))$. 
However, it can be seen that $\max(\{p_{t}(i,j)\})$ and $\min(p_{t}(i,j))$ are unstable with $K$. 
Meanwhile, the $\operatorname{avg}(p_{t}(i,j))$ and $Q_{t}(p_{k}(i,j))$ keeps a stable performance. 
A thought goes that $\min(p_{t}(i,j))$ and $\max(p_{t}(i,j))$ are easily affected by singular points with an outlier density: some high-confidence points may be diminished by $\min(p_{t}(i,j))$, and incorrect neighbors from the other cluster may be enhanced by $\max(p_{t}(i,j))$. 
The mean and median can sufficiently use the density information to get a stable result. 

We point out that Tab.~\ref{table4} does not conflict with our \textbf{Theorem2}, considering that we forgo sufficiently fine-tuning the bandwidth $h$ for $\max(p_{t}(i,j)), \min(p_{t}(i,j))$ considering $\operatorname{avg}(p_{t}(i,j))$ is more efficient and convenient options.

\subsubsection{A Mild Density-aware Strategy}

Inspired by Tab.~\ref{table5}, a thought goes that $I(p_{i,j})$ reflects the \textit{cluster assumption}, but it does not serve as a sufficient statistic. Thus an incorrect prior at the beginning may severely mislead the training. It motivates us that it's better to take PMLP as a mild punishment that does not reject the far neighbors firmly and control the PM $d(x_{i},x_{j})$ with a comparably slight disparity. 

We choose the $I(p_{i,j}) = \operatorname{avg}(\{p_{t}(i,j),t=1,2,...,k\})$ to deploy PMLP and fine-tune the bandwidth $h$ to attenuate the influence of density information, as $h$ significantly impacts the KDE. 
Tab.~\ref{table4} presents the accuracy of the PMLP on CIFAR100 with different bandwidth $h$, $K=1$. 
When $h$ is relatively small, the density of the midpoints shows significant differentiation. Thus, the affinity matrix $ W'$ is notably influenced by the density information $I(p_{i,j})$ and efficiently diminishes the far neighbors. 
The densities become similar when the $h$ is relatively large, and the $ W'$ mildly obsoletes the low-confidence neighbors. 
By \textbf{Theorem2}, setting $h \rightarrow \infty$ degenerates PMLP into a classical LPA~\cite{zhou2003learning,iscen2019label,zhao2022lassl}. 
When $h = \infty$, the accuracy of PMLP indeed converges to LASSL, which indicates that PMLP degenerates to classical LPA. 

\subsubsection{Effect of Neighbor Point Numbers $\textit{N}$}

For a fair comparison with our baseline LASSL, we keep the same choice of $1.5K$ neighbors in LPA. It is still a matter of whether more or fewer neighbors will bring better performance. Then we try other choices in Tab.~\ref{table6} for the ablation study. For CIFAR10 with 40 labeled samples, $1.5K = 15$ reaches the best performance over the $5$ choices. The result aligns with our intuition that fewer neighbors lack the information and too many neighbors may burden the LPA, for some may come from the other categories.

\section{Conclusion}
We first introduce \textbf{Theorem1} to prove that probability density helps to consider the \textit{cluster assumption} comprehensively. 
Then, we introduce a probability-density-aware measurement, PM. We design a new algorithm with PM, the Probability-Density-Aware Label-Propagation algorithm(PMLP), that propagates pseudo-labels based on clusters' geometric priors. 
\textbf{Theorem2} demonstrates PMLP's exceptional performance, attributing it to integrating density information: LPA relying on traditional distance measurements is a particular case of PMLP.
Extensive experiments are designed to show PMLP's superiority. 

\section{Acknowledgements}
The research is supported by the National Key R\&D Program of China (Grant No. 2023YFA1008700 and 2023YFA1008703), the National Natural Science Foundation of China (NO. 62102151), the Open Research Fund of Key Laboratory of Advanced Theory and Application in Statistics and Data Science, Ministry of Education (KLATASDS2305), the Fundamental Research Funds for the Central Universities. 

\bibliography{aaai25}

\newpage

\appendix

\section{Appendix}

\subsection{Proof: Theorem1}

\label{appA}

\textbf{Definition1.}\footnote{Vershynin, R. 2018. High-dimensional probability: An introduction with applications in data science. In Vol. 47. Cambridge university press.}
\textit{A zero-mean random variable X that satisfies}
\begin{equation*}
 P(|X| \geq t) \leq 2 \exp(- t^{2} / K^{2})   \label{eq1}
\end{equation*}
with any $t>0, K\in R$ is called a sub-gaussian random variable.
The sub-gaussian norm of $X$,denoted as $||X||_{\psi_{2}}$, is defined as: with any $t>0$,
\begin{equation*}
||X||_{\psi_{2}} = \inf\{ t\geq0: \exp(X^{2}/t^{2}) \leq 2 \}.
\end{equation*}

\textbf{Definition2.}\footnote{Vershynin, R. 2018. High-dimensional probability: An introduction with applications in data science. In Vol. 47. Cambridge university press.}
\textit{A random vector $X$ in $R^{d}$ is called sub-gaussian vector if the one-dimensional marginals $<X,x>$ are sub-gaussian random variables for all $x\in R^{d}$. The sub-gaussian norm of $X$ is defined as:}
\begin{equation*}
\label{1}
||X||_{\psi_{2}} = sup_{x\in S^{d-1}}||<X,x>||_{\psi_{2}}.
\end{equation*}
$S^{d-1}$ is the unit sphere with $d-1$ dimension.

\textbf{Lemma1.}
\textit{Without loss of generality, assume $X \in R^d$ is a sub-gaussian random vector with mean $\mu$ and $||X||_{\psi_{2}}\leq K$. 
Then for $t \in R^+$, with a constant $C,c$, there is:}
\begin{equation*}
P\{||X||_{2} \geq CK\sqrt{d} + t\} \leq exp(-\frac{ct^{2}}{K^2}),
\end{equation*}
\textit{Proof:}

Without loss of generality(w.l.o.g), suppose $X'$ is zero-mean gaussian random vector with $||X'||_{\psi_{2}}\leq K$.
By \textbf{6.3.5}~\cite{hdp2018}, we have: 
\begin{equation*}
P\{||BX'||_{2} \geq CK||B||_F + t\} \leq exp(-\frac{ct^{2}}{K^2||B||^2}),
\end{equation*}
where $B \in R^{m \times d}$, $||\cdot||_F$ is the Frobenius norm, $||\cdot||_2$ represents the maximum of singular value. 

Specifically suppose $B=I$, where $I \in R^{d \times d}$ is the identify matrix, there is:
\begin{equation*}
P\{||X'||_{2} \geq CK\sqrt{d} + t\} \leq exp(-\frac{ct^{2}}{K^2}),
\end{equation*}
Change the zero-mean random vector $X'$ to $X-u$ and complete the proof.

%%%%%%%%%%%%%%%%%%%%%%%%%%%%%%%%%%%%%%%%%%%%%%%%%%%%%%%%%%%%%%%%
\textbf{Lemma2.}
\textit{Assume two independent sub-gaussian random vectors $X_{1}, X_{2} \in R^d$ with $||X_{1}||_{\psi_{2}}\leq K_1$, $||X_{2}||_{\psi_{2}}\leq K_2$ and corresponding constants $C_1,C_2,c_1,c_2$, means $\mu_1, \mu_2$. Denote $x_{1} \in X_{1}$, $x_{2} \in X_{2}$ are sampled from $X_1,X_2$. 
Suppose $\gamma(x_{1},x_{2})$ is a continuous finite curve from $x_{1}$ to $x_{2}$. 
Denote set of $\gamma(x_{1},x_{2})$ as $\Gamma(x_{1},x_{2})$. 
Define set $A$:
\begin{equation*}
 A=\{x:||x-\mu_1||_2 \geq C_1K_1\sqrt{d} + t_1\},
\end{equation*}
define set $B$:
\begin{equation*}
 B=\{x: ||x-\mu_2||_2 \geq C_2K_2\sqrt{d} + t_2\},
\end{equation*}
with $t_1,t_2 \in R^+$. 
Assume $X_1.X_2$ sufficiently disperse with each other, satisfy: 
\begin{equation*}
    ||\mu_1-\mu_2||_2 > C_1K_1\sqrt{d} + C_2K_2\sqrt{d} +t_1 +t_2.
\end{equation*}
Then with probability $(1-exp(-\frac{c_1t^{2}}{K^2}))(1-exp(-\frac{c_2t^{2}}{K^2}))$, we have the fact:
\begin{equation*}
(A\cap B) \cap \gamma(x_{1},x_{2}) \neq \emptyset.
\end{equation*} }

\textit{Proof:} 

We first prove $A^c \cap B^c = \emptyset$: 
if there exist $x' \in A^c \cap B^c$, with triangle inequality, we have:
\begin{equation*}
    ||\mu_1-\mu_2||_2 \leq ||x'-\mu_2||_2 +||x'-\mu_1||_2,
\end{equation*}
\begin{equation*}
    ||x'-\mu_2||_2 + ||x'-\mu_1||_2 \leq C_1K_1\sqrt{d} + C_2K_2\sqrt{d} +t_1 +t_2,
\end{equation*}
which contradicts our assumption. 

Suppose $x_1 \in A^c,x_2 \in B^c$. 
As $\gamma(x_{1},x_{2})$ is a continuous finite curve from $x_1$ to $x_2$, there exists $x' \in \gamma(x_{1},x_{2})$ satisfies:  
\begin{equation*}
 \{x' \in \gamma(x_{1},x_{2}) , x' \in A \cap B\}.
 \end{equation*}
Then we have:
\begin{equation*}
 \{x':||x'-\mu_1||_2 \geq C_1K_1\sqrt{d}, ||x'-\mu_2||_2 \geq C_2K_2\sqrt{d}\}.
\end{equation*}
WIth the assumption $x_1 \in A^c,x_2 \in B^c$, we conclude: 
\begin{equation*}
\{(A\cap B) \cap \gamma(x_{1},x_{2})\neq \emptyset\} = \{x_1 \in A^c, x_2 \in B^c\}.
\end{equation*} 
With \textbf{Lemma1}, we have:
\begin{equation*}
P(\{x_1 \in A^c, x_2 \in B^c\}) = P(\{x_1 \in A^c, x_2 \in B^c) \neq \emptyset\}) .
\end{equation*}
\begin{equation*}
P(\{x_1 \in A^c, x_2 \in B^c) \neq \emptyset\}) \geq (1-exp(-\frac{c_1t_1^{2}}{K_1^2}))(1-exp(-\frac{c_2t_2^{2}}{K_2^2})).
\end{equation*}
Complete the proof.

%%%%%%%%%%%%%%%%%%%%%%%%%%%%%%%%%%%%%%%%%%%%%%%%%%%%%%%%%%%%%%%%
\textbf{Lemma3.}
Set $A,B$ are sets defined with parameter $t_1, t_2$ as \textbf{Lemma2}. Suppose we sample $N_1$ points from $X_1$ and $N_2$ points from $X_2$. 
Take $t_1<t_3 \in R$, $t_2<t_4 \in R$ which satisfies:
\begin{equation*}
    K(\frac{||t_3 - t_1||_2^2}{h})<\tau,~~ K(\frac{||t_4 - t_2||_2^2}{h})<\tau
\end{equation*}
with an exponential density kernel $K(\cdot)$, bandwidth $h$ and threshold $\tau$. 
Assume $X_1,X_2$ sufficiently disperse with each other:
\begin{equation*}
    ||\mu_1-\mu_2||_2 > C_1K_1\sqrt{d} + C_2K_2\sqrt{d} +t_3 +t_4.
\end{equation*}
%Then, define set $A',B'$ with $t_3, t_4$ as \textbf{Lemma2}. For any $x \in A' \cap B'$, calculate the probability density $p(x)$ with KDE. With $N_1=N_2 \rightarrow \infty$, 
%we conclude that there exists $x \in \gamma(x_1,x_2)$ satisfies $p(x)<\tau (1-a_1{K^2})(1-a_2) +  a_1a_2$ with probability $(1-a_3)(1-a_4)$, here we simplify $exp(-\frac{c_it_i^{2}}{K_i^2})$ as $a_i,~~~i=1,2$.
Then, define set $A',B'$ with $t_3, t_4$ as \textbf{Lemma2}. For any $x \in A' \cap B'$, calculate the probability density $p(x)$ with KDE. 
With $N_1=N_2 \rightarrow \infty$, we conclude: 
\begin{equation*}
 p(x)<\tau (1-\frac{1}{2}(a_1+a_2)) +  \frac{1}{2}(a_1+a_2),   
\end{equation*}
 and $\{\gamma(x_1,x_2) \cap x \in A' \cap B' \} \neq \emptyset$ with probability $(1-a_3)(1-a_4)$. 

Here we simplify $a_1, a_3 = exp(-\frac{c_1t_1^{2}}{K_1^1}), exp(-\frac{c_1t_3^{2}}{K_1^2})$; $a_2, a_4 = exp(-\frac{c_2t_2^{2}}{K_2^1}), exp(-\frac{c_2t_4^{2}}{K_2^2})$.

\textit{Proof:} 

Easy to prove that sub-gaussian vectors $X_1.X_2$ satisfies  
\begin{equation*}
    E(|X_i|) < \infty,   ~~~~i=1,2.
\end{equation*}
Thus $X_1, X_2$ satisfy the Law of Large Numbers.
Suppose $N_1,N_2 \rightarrow \infty$, denote $N(A)$ as the number of elements in set $A$, we have:
\begin{equation*}
    \frac{N(A)}{N_1} = exp(-\frac{c_1t_1^{2}}{K_1^2}), \frac{N(B)}{N_2} = exp(-\frac{c_2t_2^{2}}{K_2^2}).
\end{equation*}
With our assumption, for any $x \in A' \cap B'$, $x_i \in A^c \cup B^c$, we have:
\begin{equation*}
    ||x-x_{i}||_{2}^{2}\geq min(||t_3-t_1||_{2}^{2}, ||t_4-t_2||_{2}^{2}).
\end{equation*}
Then we have:
\begin{equation*}
    K(\frac{||x-x_{i}||_{2}^{2}}{h}) \leq K(\frac{||t_3-t_1||_{2}^{2}}{h}) \leq \tau, ~~~or
\end{equation*}
\begin{equation*}
    K(\frac{||x-x_{i}||_{2}^{2}}{h}) \leq K(\frac{||t_4-t_2||_{2}^{2}}{h}) \leq \tau.
\end{equation*}
For $x\in (A'\cap B')$, with $N_1,N_2 \rightarrow \infty$, calculate $p(x)$ with KDE:
\begin{eqnarray*}
    p(x) &=& \frac{1}{N_1+N_2} \Sigma^{N_1+N_2}_{i=1} K(\frac{||x-x_{i}||_{2}^{2}}{h})
    \\&\leq&
    \frac{\Sigma^{(x_i \in A^c \cup B^c)} K(\frac{||x-x_{i}||_{2}^{2}}{h})}{N_1+N_2} + \frac{\Sigma^{(x_i \in A \cap B)}K(\frac{||x-x_{i}||_{2}^{2}}{h})}{N_1+N_2}
    \\&\leq& \tau P(A^c \cup B^c) + P(A \cap B)
    \\&\leq& \tau (1-\frac{1}{2}(a_1+a_2)) +  \frac{1}{2}(a_1+a_2),
\end{eqnarray*}
here we simplify $exp(-\frac{c_it_i^{2}}{K_i^2})$ as $a_i,~~~i=1,2$.

A similar proof as in \textbf{Lemma2}, we have:
\begin{equation*}
\{(A'\cap B') \cap \gamma(x_{1},x_{2})\neq \emptyset\} = \{x_1 \in (A')^c, x_2 \in (B')^c\}.
\end{equation*}
With the conclusion of \textbf{Lemma2}, we have:
\begin{equation*}
P((A'\cap B') \cap \gamma(x_{1},x_{2}) \neq \emptyset) \geq (1-a_3)(1-a_4).
\end{equation*}
Here we simplify $a_3,a_4$ as in the definition. Then complete the proof.

%%%%%%%%%%%%%%%%%%%%%%%%%%%%%%%%%%%%%%%%%%%%%%%%%%%%%%%%%%%%%%%%
\textbf{Theorem1}
\textit{Suppose two sub-gaussian random vectors $X_1, X_2$ with mean $\mu_1,\mu_2$. Denote $x_1 , x_2$ randomly sampled from $X_1,X_2$, and $\gamma(x_1,x_2)$ is a continuous finite path from $x_1$ to $x_2$. For any $\tau \in R$. Denote $p(x)$ is the probability density of point $x$ calculated with KDE, define event:
\begin{equation*}
   C_\tau:\{\exists x \in \gamma(x_1,x_2), p(x)\leq \tau\}. 
\end{equation*}
 Then with $||\mu_1-\mu_2||_2^2 \rightarrow \infty$, we have: 
\begin{equation*}
    P(C_\tau) \rightarrow 1.
\end{equation*}
}
\textit{Proof:}
From \textbf{Lemma3}, we have known with probability:
\begin{equation*}
 (1-exp(-\frac{c_3t_1^{2}}{K^2})))(1-exp(-\frac{c_4t_2^{2}}{K^2})),
\end{equation*}
there $\exists x \in \gamma(x_1,x_2)$, its probability density $p(x)$ from KDE satisfies:
\begin{equation*}
    p(x) \leq \tau (1-\frac{1}{2}(a_1+a_2)) +  \frac{1}{2}(a_1+a_2),
\end{equation*}
where $a_i = exp(-\frac{c_it^{2}}{K^2}) ,~~~i=1,2$. 
Considering that $t_3,t_4$ satisfies:
\begin{equation*}
    ||\mu_1-\mu_2||_2 > C_1K_1\sqrt{d} + C_2K_2\sqrt{d} +t_3 +t_4,
\end{equation*}
when $||\mu_1-\mu_2||_2 \rightarrow \infty$, we can take $t_3, t_4 \rightarrow \infty$. 
Considering $t_1,t_2$ satisfies:
\begin{equation*}
    K(\frac{||t_3 - t_1||_2^2}{h})<\tau,~~ K(\frac{||t_4 - t_2||_2^2}{h})<\tau,
\end{equation*}
easy to prove that we can take $t_1,t_2 \rightarrow \infty$. 
Then we have:
\begin{equation*}
 \lim_{t_3,t_4 \rightarrow \infty}(1-exp(-\frac{c_3t_1^{2}}{K_1^2}))(1-exp(-\frac{c_4t_2^{2}}{K_2^2})) \rightarrow 1,
\end{equation*}
$a_i = \lim_{t_i \rightarrow \infty}exp(-\frac{c_it_i^{2}}{K_i^2}) \rightarrow 0 ,~~~i=1,2$. 

Then we have the fact: when $||\mu_1-\mu_2||_2 \rightarrow \infty$, with probability 1, we have the fact:
\begin{equation*}
\{\exists x \in \gamma(x_1,x_2), ~~~~ satisfies ~~p(x) \leq \tau\}.    
\end{equation*}

Then complete the proof.

%%%%%%%%%%%%%%%%%%%%%%%%%%%%%%%%%%%%%%%%%%%%%%%%%%%%%%%%%%%%%%%%
\subsection{Proof: Corollary1}
\textbf{Corollary1}
\textit{Specifically take $\gamma(x_1,x_2)$ in \textbf{Theorem1} as a straight line and $x\in\gamma(x_1,x_2)$ distributes uniformly. For $\forall \tau \in R$, with sufficient large $||\mu_1-\mu_2||_2^2$, we have:} 
\begin{equation*}
    \lim_{||\mu_1-\mu_2||_2 \rightarrow \infty}P(\{x:p(x) \leq \tau\ , x \in \gamma(x_1,x_2)\}) \rightarrow 1. 
\end{equation*}

\textit{Proof:}

W.l.o.g, take $\forall M < 0.5$, take $|\mu_i| \rightarrow \infty , i=1,2$ and $||\mu_1-\mu_2||_2 \rightarrow \infty$.  
Specifically take $|t_i| = M|\mu_i|, i=1,2$. It can be easily proved that we can further choose proper $t_3,t_4$, so that $t_i,~~i=1,2,3,4$ satisfies the assumptions in \textbf{Lemma2} and \textbf{Lemma3}:
\begin{equation*}
    ||\mu_1-\mu_2||_2 > C_1K_1\sqrt{d} + C_2K_2\sqrt{d} +t_1 +t_2,
\end{equation*}
\begin{equation*}
    ||\mu_1-\mu_2||_2 > C_1K_1\sqrt{d} + C_2K_2\sqrt{d} +t_3 +t_4.
\end{equation*}

We denote the above assumptions with $M$ as $\Omega_M$.
Then for $x_i$ sampled from $X_i$, with \textbf{Lemma1}, we have:
\begin{equation*}
P(x_i:\lim_{||\mu_1-\mu_2||_2^2 \rightarrow \infty}||x_i-\mu_i||_2 \leq C_iK_i\sqrt{d} + t_i) \rightarrow 1. 
\end{equation*}

Then as the setting in \textbf{Lemma2} and \textbf{Lemma3}, define $A,B,A',B'$ with corresponding $t_1,t_2,t_3,t_4$. 
Denote $l'(x_1,x_2), l''(x_1,x_2)$:
\begin{equation*}
 l'(x_1,x_2) = \{x\in \gamma(x_1,x_2), x\in A\cup B\},
\end{equation*}
\begin{equation*}
 l''(x_1,x_2) = \{x\in \gamma(x_1,x_2), x\in A'\cup B'\}. 
\end{equation*}
Then with our assumption $\Omega_M$ we have:
\begin{equation*}
  \frac{||l'(x_1,x_2)||_2}{||\gamma(x_1,x_2)||_2} \leq   \frac{(1-2M)||\mu_1-\mu_2||_2}{(1+2M)||\mu_1-\mu_2||_2},
\end{equation*}
\begin{equation*}
  \frac{(1-2M)||\mu_1-\mu_2||_2}{(1+2M)||\mu_1-\mu_2|||_2}  \rightarrow \frac{1-2M}{1+2M}
\end{equation*}
with probability 1.

With $\Omega_M$, from \textbf{Lemma3}, it can be calculated that $||t_i-t_j||_2^2 \leq c(\tau)$, where $\{i=1,j=3\}$ or $\{i=2,j=4\}$, $c(\tau) \in R$ is a constant with respect to $\tau$,$h$ and $K(\cdot)$. Then we have:
\begin{equation*}
\lim_{t_1,t_2 \rightarrow \infty}\frac{||l''(x_1,x_2)||_2}{||l'(x_1,x_2)||_2} \rightarrow 1.
\end{equation*}
Then we have:
\begin{equation*}
\lim_{t_1,t_2 \rightarrow \infty}\frac{||l''(x_1,x_2)||_2}{||l(x_1,x_2)||_2} \rightarrow \frac{1-2M}{1+2M},
\end{equation*}
for $\forall M\leq 0.5$. 

W.l.o.g, take $M \rightarrow 0$, we have:
\begin{equation*}
\lim_{t_1,t_2 \rightarrow \infty}\frac{||l''(x_1,x_2)||_2}{||l(x_1,x_2)||_2} \rightarrow \frac{1-2M}{1+2M}=1,
\end{equation*}
Considering $x \in \gamma(x_{1},x_{2}) = l(x_1,x_2)$ is uniformly distributed, we have:
\begin{equation*}
\lim_{||\mu_1-\mu_2||_2^2 \rightarrow \infty}\frac{P(x \in \gamma(x_{1},x_{2}))}{P(x \in \gamma'(x_{1},x_{2}))} \rightarrow 1. 
\end{equation*}
With \textbf{lemma3} and \textbf{Theorem1}, we have the fact: the event, $\{x \in \gamma(x_1,x_2)$ , $p(x) \leq \tau\}$, holds with probability 1.

Complete the proof.

%%%%%%%%%%%%%%%%%%%%%%%%%%%%%%%%%%%%%%%%%%%%%%%%%%%%%%%%%%%%%%%%
\subsection{Proof: Theorem2}

\textbf{Lemma4.} \textit{Suppose pseudo-label $\widehat Y^{LP}$ is propagated with an affinity matrix $W$, and $\widehat Y'^{LP}$ is propagated by an affinity matrix $W'$:}
\begin{equation*}
    \widehat Y^{LP} = (I-\alpha D^{-\frac{1}{2}} W D^{-\frac{1}{2}})^{-1}Y^{LP,\mathrm{high}},
\end{equation*}
\begin{equation*}
    \widehat Y'^{LP} = (I-\alpha D'^{-\frac{1}{2}} W' D'^{-\frac{1}{2}})^{-1}Y^{LP,\mathrm{high}}.
\end{equation*}
\textit{Moreover, suppose $W=kW', k\in R$,  and $D$ and $D'$  are two diagonal matrices with their $(i, i)$-th elements being the sum of the $i$-th row of $W$ and $W'$ respectiely.
Then we have
}
\begin{equation*}
    \widehat Y'^{LP} = \widehat Y^{LP}
\end{equation*}
\textit{Proof:}
\begin{eqnarray*}
     \widehat Y'^{LP} &=& (I-\alpha D'^{-\frac{1}{2}} W' D'^{-\frac{1}{2}})^{-1}Y^{LP,\mathrm{high}}  
     \\&=&(I-\alpha (kD)^{-\frac{1}{2}} (kW) (kD)^{-\frac{1}{2}})^{-1}Y^{LP,\mathrm{high}}
     \\&=&(I-\alpha (\frac{1}{\sqrt{k}}D^{-\frac{1}{2}}) (kW) (\frac{1}{\sqrt{k}}D^{-\frac{1}{2}}))^{-1}Y^{LP,\mathrm{high}}
     \\&=&(I-\alpha D^{-\frac{1}{2}} W D^{-\frac{1}{2}})^{-1}Y^{LP,\mathrm{high}}
     \\&=&\widehat Y^{LP}
     \\&&
\end{eqnarray*}
The proof is completed.

%%%%%%%%%%%%%%%%%%%%%%%%%%%%%%%%%%%%%%%%%%%%%%%%%%%%%%%%%%%%%%%%
\textbf{Theorem2.}
\textit{
Denote the model parameter in one iteration as $\theta$.
%The encoder is $h(\cdot)$, and the predictor is $p(\cdot)$. 
And let $\mathcal{L}_{U}$ be the unsupervised loss between predictions and pseudo-labels.
%Denote $O^{LP}$ as the set of features and $Y^{P}$ as the prediction vector.
%The $i$-th element of $Y^{P}$ is then $y_{i}^{P} = f\cdot h(x_{i})$.
Denote $Y^{P}$ as the set of predictions, $\widehat Y^{*LP}$ and $\widehat Y^{LP}$ be the pseudo-labels generated by PMLP and traditional LPA with any 
 $D(\cdot,\cdot)$. 
Suppose we calculate the probability density by KDE with exponential kernel and bandwidth $h$. 
Then $\mathcal{L}_{U}$ satisfies:}
\begin{equation*}
\min_{\theta \in \Theta, h\in R}\mathcal{L}_{U}(Y^{P}, \widehat Y^{*LP}) \leq \min_{\theta \in \Theta}\mathcal{L}_{U}(Y^{P}, \widehat Y^{LP}).
\end{equation*}
\textit{Proof:}
To prove \textbf{Theorem2}, it sufficies to proof that: 
   \begin{equation*}
   \min_{\theta \in \Theta}L_{U}(Y^{P},\widehat Y^{LP}) = \min_{\theta \in \Theta, h=h^{*}}L_{U}(Y^{P},\widehat Y^{*LP}).
   \end{equation*}
Then there is:
\begin{equation*}
   \min_{\theta \in \Theta, h\in R}L_{U}(Y^{P},\widehat Y^{*LP}) \leq \min_{\theta \in \Theta, h=h^{*}}L_{U}(Y^{P},\widehat Y^{*LP}),
   \end{equation*}
\begin{equation*}
   \min_{\theta \in \Theta, h=h^{*}}L_{U}(Y^{P},\widehat Y^{*LP}) = \min_{\theta \in \Theta}L_{U}(Y^{P},\widehat Y^{LP}),
   \end{equation*}
which can complete the proof. 

Denote $\widehat Y^{*LP}$ with $h = h^{*}$ as $\widehat Y^{*LP}(h^{*})$, it suffices to prove that $\widehat Y^{*LP}(h^{*}) = \widehat Y^{LP}$. 

Denote the set of high-confidence predictions as $Y^{LP,\mathrm{high}}$, the set of low-confidence predictions as $Y^{LP, \mathrm{low}}$, the affinity matrix $W$, with hyper-parameter $\alpha$, $\eta$ in pseudo-labeling. Then there is:
\begin{equation*}
   \widehat Y^{LP} = \eta (I-\alpha D^{-\frac{1}{2}} W D^{-\frac{1}{2}})^{-1}Y^{LP,\mathrm{high}} + (1 - \eta)(Y^{LP,\mathrm{low}}),
\end{equation*}
note that PMLP's affinity matrix $W'$ is adjusted by density information $I(p_{i,j})$ corresponding to bandwidth $h$, then there is:
\begin{footnotesize} 
\small
\begin{eqnarray*}
 \widehat Y^{*LP}(h) = \eta (I-\alpha D'(h)^{-\frac{1}{2}}W'(h)D'(h)^{-\frac{1}{2}})^{-1}Y^{LP,\mathrm{high}}
   + (1 - \eta)Y^{LP,\mathrm{low}}.
\end{eqnarray*}
\end{footnotesize} 

Then to prove $\widehat Y^{*LP}(h^{*}) = \widehat Y^{LP}$, it suffices to show that 
 \begin{equation*}
   D'(h^{*})^{-\frac{1}{2}}W'(h^{*})D'(h^{*})^{-\frac{1}{2}} = D^{-\frac{1}{2}}WD^{-\frac{1}{2}}, ~~~~~~   \mathrm{with} ~~ h^{*} \in R.
\end{equation*}

Denote that 
\begin{equation*}  
W_{i,j} = \left\{  
\begin{alignedat}{2}  
&0,                    & \quad &\text{if } i=j, \\  
&D(o_{i}^{LP},o_{j}^{LP}),                    & \quad &\text{if } i \neq j,  \\  
\end{alignedat}  
\right.  
\end{equation*} 
 
\begin{equation*}  
W'_{i,j}(h) = \left\{  
\begin{alignedat}{2}  
&0,                    & \quad &\text{if } i=j, \\  
&I(p_{i,j})(h) D(o_{i}^{LP},o_{j}^{LP}),                   & \quad &\text{if } i \neq j.  \\  
\end{alignedat}  
\right.
\end{equation*}

Denote $p_{i,j}^{l}(h), ~~l \in \{1,2,...,k\}$ represents the density of the point $o_{i,j}^{l},~~l \in \{1,2,...,k\}$, where $o_{i,j}^{l}$ represents the $l$-th-equal division points between two features $o_{i}^{LP}, o_{j}^{LP} \in O^{LP}$. 
To calculate the density $\{p_{i,j}^{1},...,p_{i,j}^{k}\}$, for each point $o_{i,j}^{l}$, we select $n$ points in its neighbor: $\{o_{i,j}^{l,1},...,o_{i,j}^{l,n}\}$, and calculate $p_{i,j}^{l}$ with KDE whose bandwidth is set as $h$: 
\begin{equation*}
    p_{i,j}^{l}(h) = \frac{1}{nh} \Sigma^{n}_{m=1} K(\frac{||o_{i,j}^{l,m} - o_{i,j}^{l}||_{2}^{2}}{h}),
\end{equation*}

Without loss of generality, we take $I(p_{i,j})$ as the average $k$ equal division points' density as $p_{i,j}$:
\begin{equation*}
    p_{i,j}(h) = \frac{1}{nhk} \Sigma^{k}_{l=1}\Sigma^{n}_{m=1} K(\frac{||o_{i,j}^{l,m} - o_{i,j}^{l}||_{2}^{2}}{h}),
\end{equation*}
 where $K(\cdot)$ is a Exponential kernel satisfies:
\begin{equation*}
    K(x) = exp(-x).
\end{equation*}
With \textbf{Lemma3}, it equals to update $W$ with:
\begin{equation*}  
W'_{i,j}(h) = \left\{  
\begin{alignedat}{2}  
&0,                    & \quad &\text{if } i=j, \\  
&p'_{i,j}(h) D(o_{i}^{LP},o_{j}^{LP}).                   & \quad &\text{if } i \neq j,  \\  
\end{alignedat}  
\right.
\end{equation*} 
to get a same $\widehat Y^{*LP}(h)$, where $p'_{i,j}$ satisfies:
\begin{equation*}
    p'_{i,j}(h) = \frac{1}{nk}\Sigma^{k}_{l=1}\Sigma^{n}_{m=1} K(\frac{o_{i,j}^{l,m} - o_{i,j}^{l}}{h}),
\end{equation*}
When $h \rightarrow \infty$, there is:
\begin{equation*}
    \lim_{h \rightarrow \infty} p'_{i,j}(h) =
    \lim_{h \rightarrow \infty}\frac{1}{nk}\Sigma^{k}_{l=1}\Sigma^{n}_{m=1} K(\frac{o_{i,j}^{l,m} - o_{i,j}^{l}}{h}) = 1.
\end{equation*}
Then we have 
\begin{equation*}
    \lim_{h \rightarrow \infty} p'_{i,j}(h) = 1,~~~~~ \lim_{h \rightarrow \infty}W'(h) = W.
\end{equation*}
That is denote $h^{*}= \infty$, there is:
\begin{equation*}
    \widehat Y^{*LP}(h^{*}) = \widehat Y^{LP}, 
\end{equation*}
\begin{equation*}
    \min_{\theta \in \Theta,h\in R}L_{U}(Y^{LP},\widehat Y^{*LP}) \leq \min_{\theta \in \Theta,h^{*}=\infty}L_{U}(Y^{LP},\widehat Y^{*LP}(h^{*})),
\end{equation*}
\begin{equation*}
    L_{U}(Y^{LP},\widehat Y^{*LP}(h^{*})) = \min_{\theta \in \Theta}L_{U}(Y^{LP},\widehat Y^{LP}).
\end{equation*}
Complete the proof. The case when $I(p_{i,j})$ takes maximum, minimum or t-quantile between $\{p(x_{i,j}^1,...,p(x_{i,j}^k])\}$ can be proved in a similarity way.

\subsection{PMLP Generates Better Pseudo-labels: Extension}
\begin{figure*}[t!]  
  \centering  
  
  \begin{minipage}{.33\textwidth}   
    \centering  
    \includegraphics[width=\linewidth]{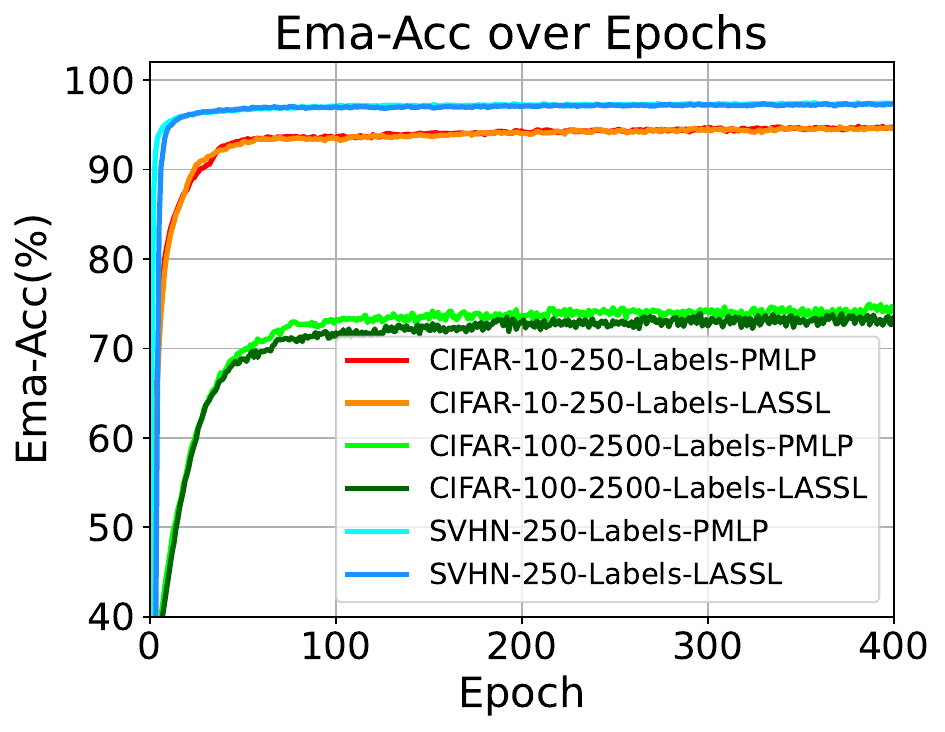}  
  \end{minipage}   
  \begin{minipage}{.33\textwidth}   
    \centering  
    \includegraphics[width=\linewidth]{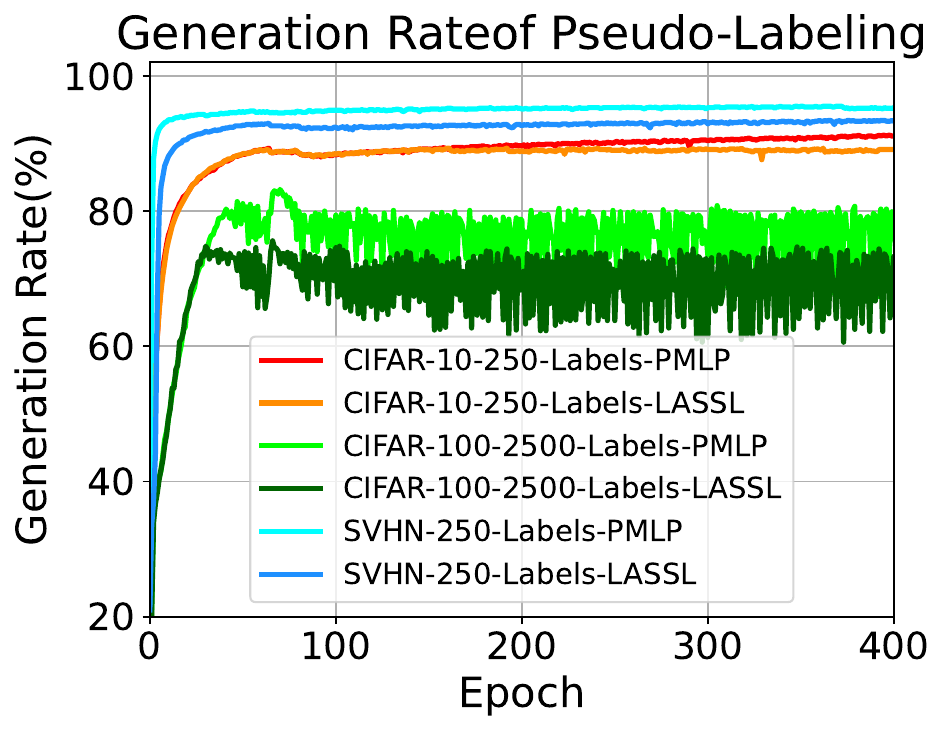}  
  \end{minipage}  
    \begin{minipage}{.33\textwidth}
    \centering
    \includegraphics[width=\linewidth]{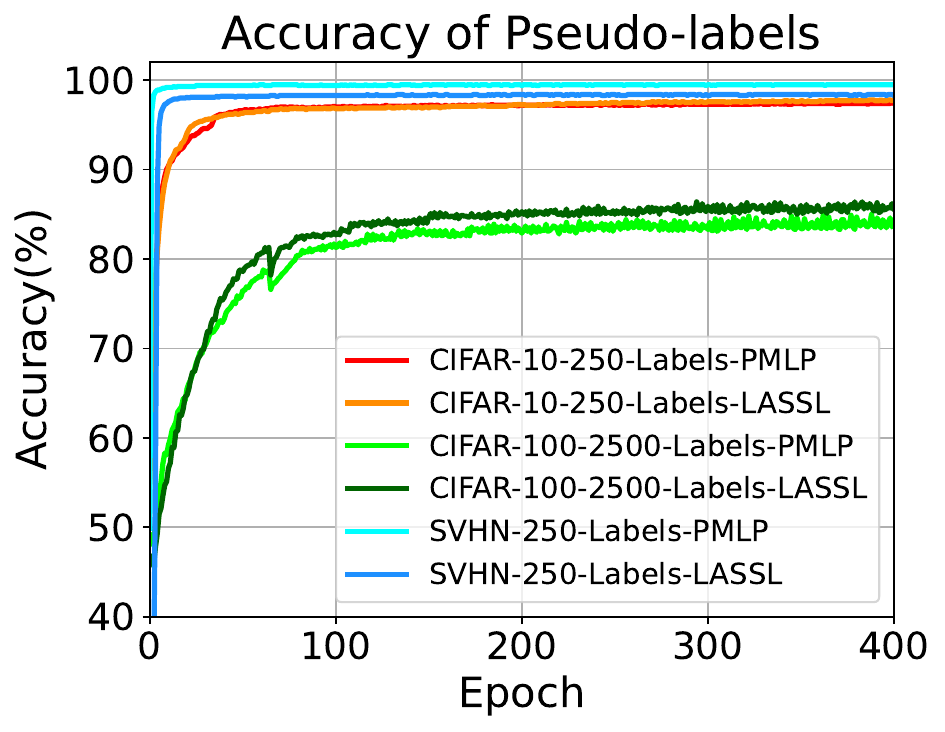}
  \end{minipage}
  \caption{
  The accuracy, rate of high-quality predictions, and accuracy of pseudo-labels on CIFAR10 with 250 labeled data, CIFAR100 with 2500 labeled data, and SVHN with 250 labeled data. It can be seen that PMLP can still produce more correct pseudo-labels, which conform to our conclusion in the main context.
  }\label{2222222}
\end{figure*}

\begin{figure*}[t!]  
  \centering    
  \begin{minipage}{.495\textwidth}   
    \centering  
    \includegraphics[width=\linewidth]{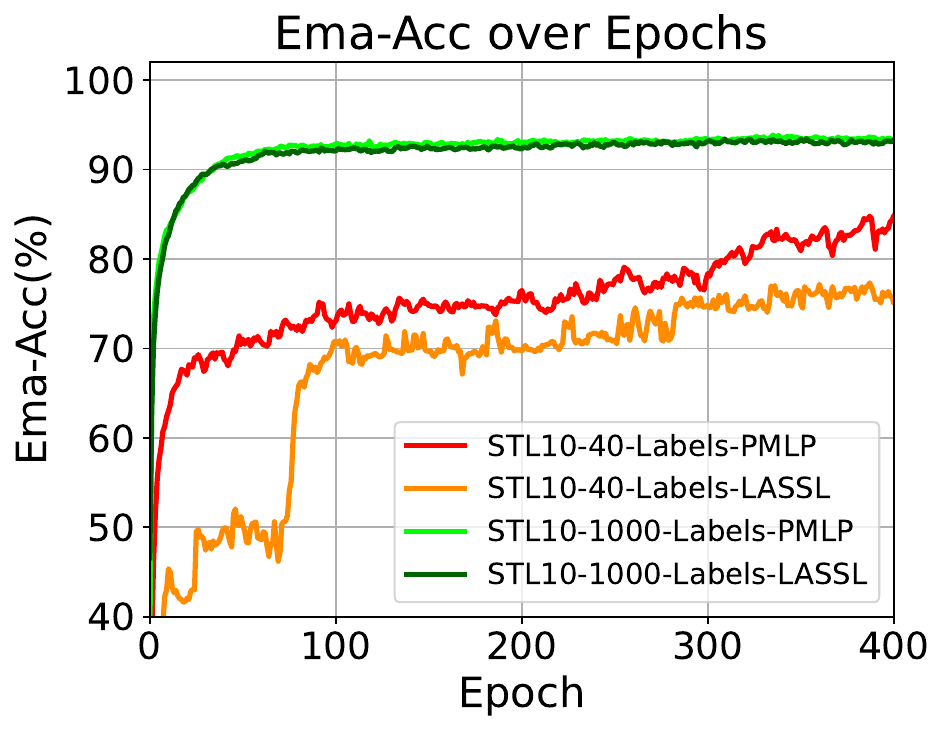}  
  \end{minipage}  
    \begin{minipage}{.495\textwidth}
    \centering
    \includegraphics[width=\linewidth]{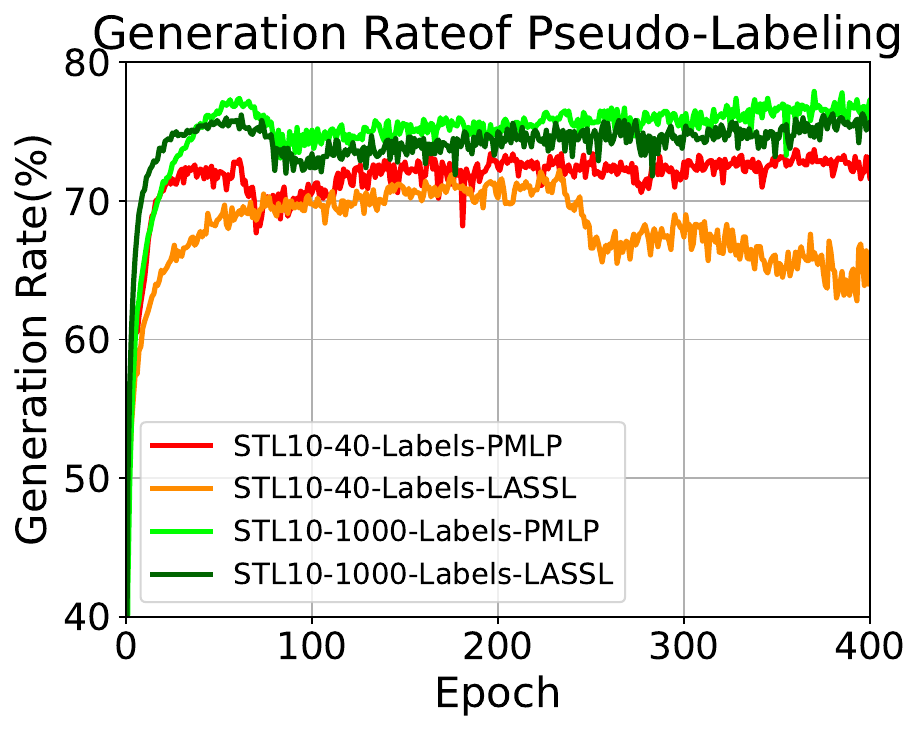}
  \end{minipage}
  \caption{
  The accuracy and rate of high-quality predictions on STL-10 with 40 and 1000 labeled data. It can be seen that PMLP can still produce more correct pseudo-labels, which conform to our conclusion in the main context.
  }\label{333333333}
\end{figure*}

We demonstrate that PMLP can generate more accurate pseudo-labels in the main context.
This section presents additional examples with different datasets to prove that PMLP can generally produce more correct pseudo-labels. 
It can be seen from Fig.~\ref{2222222} and Fig.~\ref{333333333} that with a more considerable amount of labeled data, PMLP still performs better than LPA and can generate more accurate pseudo-labels.

\subsection{GPU Based KDE: Details}

The KDE from Sklearn runs only on the CPU, which slows down PMLP's calculation. 
To accelerate PMLP, we design the GPU-based KDE:
Given the chosen point $o_{i,j}$ and the neighboring points ${o^{1}_{i,j},...,o^{n}_{i,j}}$, we calculate $p_{i,j}(h)$ on the GPU as follows:
\begin{equation*}
p_{i,j}(h) = \frac{1}{n}\sum_{m=1}^{n}K\left(\frac{||o_{i,j}^{l,m} - o_{i,j}^{l}||_{2}^{2}}{h}\right).
\end{equation*} 
In our approach, we renew a GPU-based KDE algorithm, which can be seen in the algorithm~\ref{algorithm2}. The algorithm can accelerate the KDE significantly, as seen in the main context's experiments.

\begin{algorithm}[tb] 
\caption{Quick density-aware approach}  
\label{algorithm2}
\textbf{Input:} the set of features $O^{LP}$, 
                       $\tau$, $k$,
                       the set of midpoints $\{\frac{o_{i}^{LP} + o_{j}^{LP}}{2}\}, ~~ i,j \in \{1,...,k\}$,
                       the $n$ neighbor points for different midpoints, 
                       $\{o^{1}_{i,j},...,o^{n}_{i,j}\},~~ i,j \in \{1,...,k\}$, 
                       Exponential kernel $K(\cdot)$, 
                       bandwidth $h$.
                       
\textbf{Output:} updated threshold $p_{i,j}$.
\begin{algorithmic}[1] 
\STATE Divide the set of features $O^{LP}$ into $O^{LP,high}$, $O^{LP,low}$ by threshold $\tau$;\\
\STATE For any $o_{i}^{LP} \in O^{LP,low}$, select the $k$ nearest neighbor points in $O^{LP,high}$ and calculate the midpoints $\{\frac{o_{i}^{LP} + o_{j}^{LP}}{2}\}, ~~ i,j \in \{1,...,k\}$;\\
\STATE Choose the $n$ nearest neighbor points for each $\frac{o_{i}^{LP} + o_{j}^{LP}}{2}$, get $\{o^{1}_{i,j},...,o^{n}_{i,j}\},~~ i,j \in \{1,...,k\}$;\\
\FOR {\textbf{each} $o^{m}_{i,j}$,} 
\STATE{Calculate $K(\frac{||o_{i,j}^{l,m} - o_{i,j}^{l}||_{2}^{2}}{h})$};
\ENDFOR\\
\STATE Average the $n$ $K(\frac{||o_{i,j}^{l,m} - o_{i,j}^{l}||_{2}^{2}}{h})$ with $\frac{1}{n}$, without $h$.\\
\end{algorithmic}  
\end{algorithm}

\subsection{Toward Better $I(p_{i,j})$: Extension}
We have studied which choice of $I(p_{i,j})$ is better in the main context by comparing the performance of PMLP with different $I(p_{i,j})$ on CIFAR10, 40 labeled samples. The result suggests that $avg\{p_{t}(i,j),t=1,2,...,k\}$ and $Q_t\{p_{t}(i,j),t=1,2,...,k\}$ tend to have a stable good performance and $max\{p_{t}(i,j),t=1,2,...,k\}$ or $min\{p_{t}(i,j),t=1,2,...,k\}$ is unstable.

\begin{table}[t] 
  \centering
\fontsize{7.5}{12}\selectfont
  \begin{tabular}{c|c|c|c|c}
    \toprule
    \hline
    Bandwidth  & h=5   & h=10      &h=100    &LASSL  \\
    \hline
    $max\{p_{t}(i,j)\}$      &92.72± 3.38  &95.29± 0.86  &95.06±0.61                &95.07±0.78\\
    $R_h$, density-ratio   &1.3   &1.15         &1.01    &1\\
    \hline
    $min\{p_{t}(i,j)\}$      &92.42± 2.3  &95.13± 0.81  &95.03±0.64                      &95.07±0.78\\
    $R_h$, density-ratio   &1.3   &1.15       &1.01      &1\\
    \midrule
    \bottomrule 
  \end{tabular}
  \caption{Ablation study on CIFAR10 with $40$ labeled data. We modify bandwidth $h$ to diminish the influence of density information $max\{p_{t}(i,j)\}$ and $min\{p_{t}(i,j)\}$. It can be seen $max\{p_{t}(i,j)\}$ and $min\{p_{t}(i,j)\}$ are sensitive to density information and comparably unstable compared to $avg\{p_{t}(i,j)\}$. }
  \label{1111111111}
\end{table}

In this section, we show the unstable of $max\{p_{t}(i,j),t=1,2,...,k\}$ or $min\{p_{t}(i,j),t=1,2,...,k\}$ dues to their sensitivity to neighbor's densities. 
We first define density ratio $R_h=\frac{max\{p_{t}(i,j),t=1,2,...,k\}}{min\{p_{t}(i,j),t=1,2,...,k\}}$ and estimate $R_h$ by randomly sampling 1000 pairs of $p_{t}(i,j)$ from the former 5 epoches. As discussed in the main text, $I(p_{i,j})$ is not a sufficient statistic for reflecting the \textit{cluster assumption} and $R_h$ can reflect our confidence in detecting the \textit{cluster assumption} with $max\{p_{t}(i,j),t=1,2,...,k\}$ and $min\{p_{t}(i,j),t=1,2,...,k\}$. 

In the main context, we choose the bandwidth $h=5$, with $R_h = 1.3$. When the bandwidth increases to $h=10,100$, the density ratio decreases to $1.15,1.01$ and restrains the influence of singular points. 
We deploy our experiments on the CIFAR10 dataset with 40 labeled samples, choose $k=3$, and repeat the experiments thrice. 
As can be seen in Tab.~\ref{1111111111}, with lower $R_h$, the $max\{p_{t}(i,j),t=1,2,...,k\}$ and $min\{p_{t}(i,j),t=1,2,...,k\}$ tend to get a stable good performance, which reveals that $max\{p_{t}(i,j),t=1,2,...,k\}$ and $min\{p_{t}(i,j),t=1,2,...,k\}$ are sensitive to point's density and thus can be easily affected by singular values. A large bandwidth $h$ helps reduce $R_h$ and can help get a stable result. 

It seems that decreasing $R_h$ can help to stablize the $max\{p_{t}(i,j),t=1,2,...,k\}$ and $min\{p_{t}(i,j),t=1,2,...,k\}$. However, by \textbf{theorem2}, $\lim_{h\rightarrow \infty}R_h \rightarrow 1$, which means the cost is to diminish density information $I(p_{i,j})$ and degenerating PMLP to LPA. As seen in Tab.~\ref{1111111111}, when $h=10$, $max\{p_{t}(i,j),t=1,2,...,k\}$ and $min\{p_{t}(i,j),t=1,2,...,k\}$ tend to be better. However, when $h=10$, the standard deviation of $max\{p_{t}(i,j),t=1,2,...,k\}$ and $min\{p_{t}(i,j),t=1,2,...,k\}$ is \textbf{0.86} and \textbf{0.81}, which is larger than $avg\{p_{t}(i,j),t=1,2,...,k\}$'s \textbf{0.32} in the main context. 
To this issue, we conclude that $avg\{p_{t}(i,j),t=1,2,...,k\}$ is the better choice of $I(p_{i,j})$.

\subsection{Ablgation Study: Toward The Balance with Time and Accuracy}
In PMLP, we specifically, here we take $I(P_{i,j})$ as $avg\{p_{t}(i,j),t=1,2,...,k\}$ or $Q_t\{p_{t}(i,j),t=1,2,...,k\}$, where $p_{t}(i,j)$ are chosen from the connecting line $l(x_1,x_2)$ with the equal distance. 
A natural thought goes that increasing $k$ will improve PMLP's robustness: view $I(p_{i,j})$ as a random variable, more samples help to get a stable and accurate result due to the law of large numbers. 
\begin{table*}
\centering
\renewcommand{\arraystretch}{1.2}
\begin{tabular}{c|c|c|c|c|c}
\toprule
 \hline
 \multirow{2}{*}{Number of points} & \multirow{2}{*}{Overall Average Time (s/epoch)} &\multicolumn{4}{c}{Average Time Across Different Iterations (s/epoch)} \\
\cline{3-6}
 &   & 1-256 & 257-512 & 513-768 & 769-1024 \\
\hline
baseline, LASSL & 211.41 & 255.52 & 196.10 & 196.66 & 197.35 \\ 
\cline{1-6}          
\multirow{1}{*}{PMLP, K=1} & 217.94 & 260.05 & 203.72 & 203.52 & 204.46 \\
  Relative Time Increase vs baseline        &\textbf{+3.1\%} & \textbf{+1.75\%} & \textbf{+3.9\%} & \textbf{+3.49\%} & \textbf{+3.6\%}  \\
\cline{1-6} 
  PMLP, K=3               & 248.82 & 285.31 & 236.08 & 236.65 &237.23  \\  
  Relative Time Increase vs K=1        &\textbf{+14.14\%} & \textbf{+9.72\%} & \textbf{+15.89\%} & \textbf{+16.28\%} & \textbf{+16.03\%}  \\
  Relative Time Increase vs baseline        &\textbf{+17.7\%} & \textbf{+11.66\%} & \textbf{+20.39\%} & \textbf{+20.33\%} & \textbf{+20.21\%}  \\ 
 \hline
\multirow{1}{*}{PMLP, K=5} & 310.25 & 381.09 & 307.12 & 276.55 & 276.23  \\ 
   Relative Time Increase vs K=1       &\textbf{+42.48\%} & \textbf{+46.54\%} & \textbf{+50.76\%} & \textbf{+35.88\%} & \textbf{+35.10\%}  \\
   Relative Time Increase vs baseline        &\textbf{+46.75\%} & \textbf{+49.14\%} & \textbf{+56.61\%} & \textbf{+40.62\%} & \textbf{+39.97\%}  \\

  \hline
 \bottomrule
\end{tabular}
\caption{Comparison of overall average time and average time across different training phases for different methods on STL-10 and CIFAR-10 datasets.}
\label{135}
\end{table*}
\begin{figure*}[t!]  
  \centering  
  \begin{minipage}{.33\textwidth}   
    \centering  
    \includegraphics[width=\linewidth]{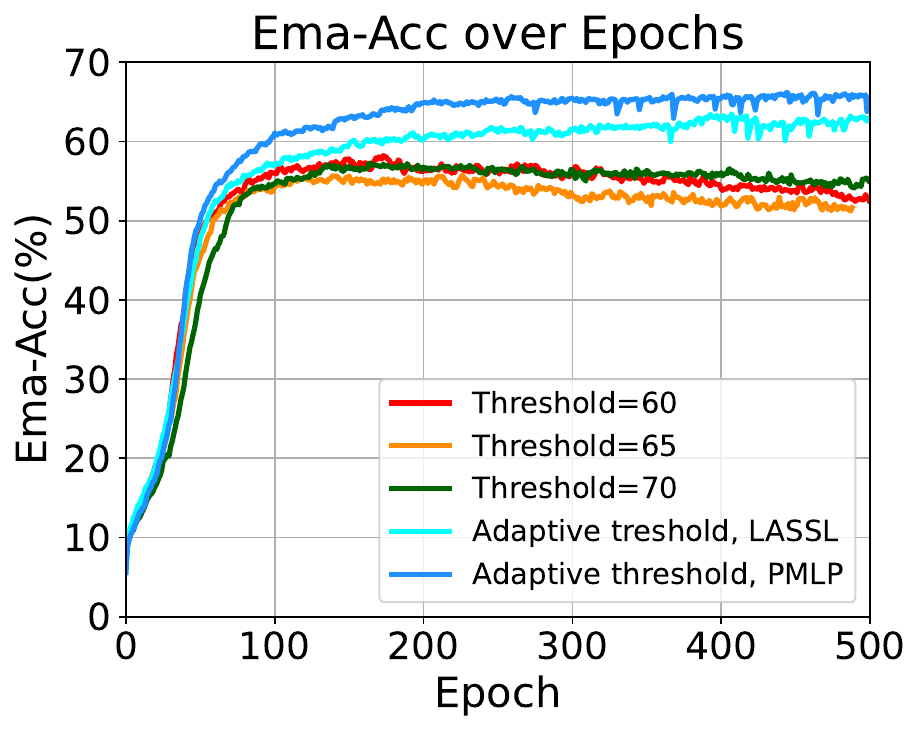}  
  \end{minipage}   
  \begin{minipage}{.33\textwidth}   
    \centering  
    \includegraphics[width=\linewidth]{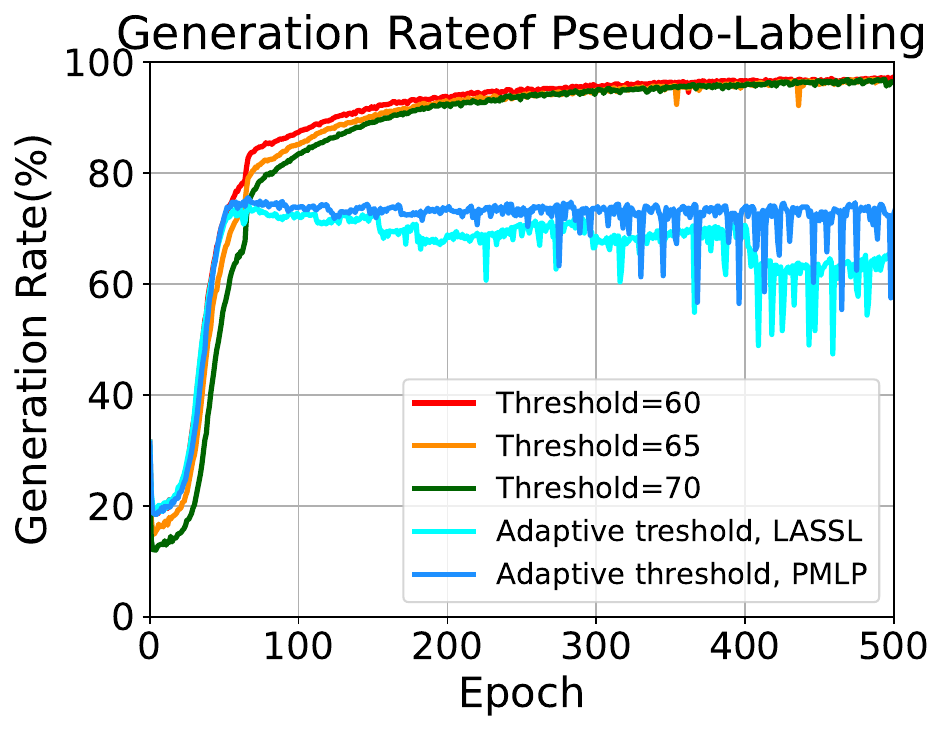}  
  \end{minipage}  
    \begin{minipage}{.33\textwidth}
    \centering
    \includegraphics[width=\linewidth]{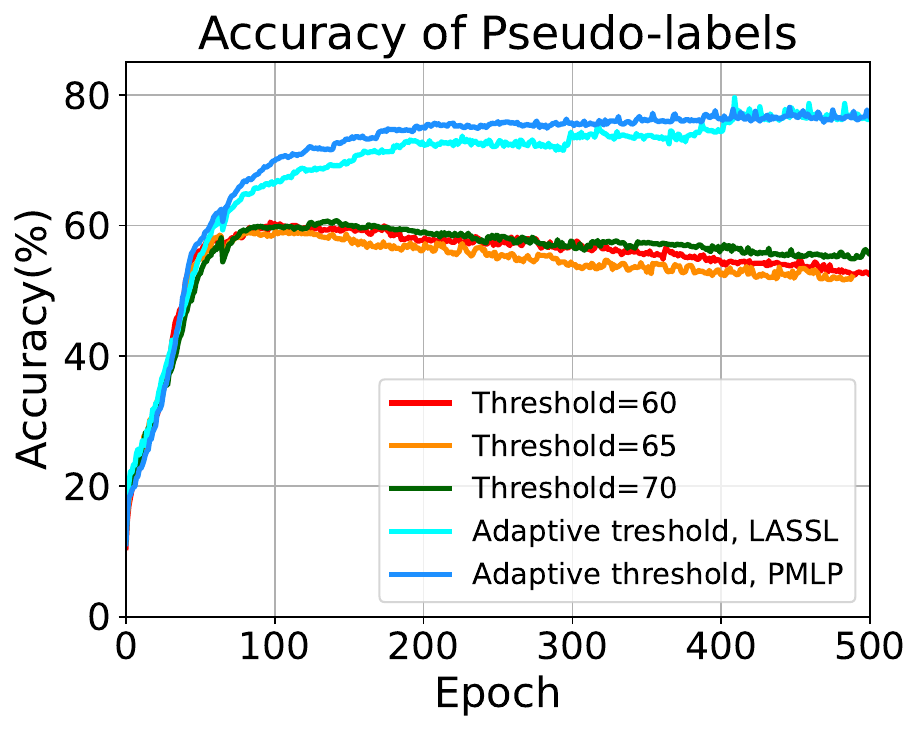}
  \end{minipage}
  \caption{
  The accuracy, rate of high-quality predictions, and correct pseudo-label ratios.
  It can be seen that LASSL with tuning strategy outperforms the LASSL with fixed $\tau$. With the same tuning strategy, PMLP outperforms the LASSL.
  }\label{fig520}
\end{figure*}
However, increasing $K$ will increase the time for KDE. In this section, we compare the time consumption of different $k$. We deploy PMLP with $K=1,3,5$ on CIFAR10 with 40 labeled data, and the $I(p_{i,j})$ is chosen as $avg\{p_{t}(i,j),t=1,2,...,k\}$. 
It can be seen in Tab.~\ref{135} that with $k$ increase, the time consumption increases obviously. From our main context, PMLP with $K=1$ only increases about $3\%$ time consumption compared with baseline LASSL, but when $K=3$, the PMLP increases about $18\%$ time consumption compared with our baseline LASSL. 
When $k=5$, the PMLP increases about $47\%$ time consumption compared with our baseline LASSL.
As the main context shows, $k=3,5$ can not bring an apparent breakthrough, and $k=1$ has significantly improved PMLP than traditional LPA. This motivates us to believe that choosing $k=1$ is enough to achieve PMLP's superior performance without adding too much calculation burden.

\subsection{An Adaptive Threshold Tuning Strategy}

In the baseline model LASSL, high-confidence pseudo-labels are selected for training, defined as $y_{i}^{LP,\mathrm{high}} = I(\max(y_{i}^{LP}) \geq \tau)y_{i}^{LP}$. The threshold for assigning a high-quality label is set to $\tau=0.95$ across all experiments on SVHN, CIFAR10, and CIFAR100 datasets. 

\begin{algorithm}[tb] 
\caption{Adaptive threshold tuning strategy}
\label{appC}
\textbf{Input:}       predictions $Y^{P}$,
                      the threshold $\tau$ from the last iteration,
                      counting number $i$,
                      training epoch $n$.
\textbf{Output:} updated threshold $\tau$.
\begin{algorithmic}[1] 
\FOR {$n = 1$ to $1024$}
\FOR{\textbf{each} iteration:}
\IF{$p_{i,j} > \tau$}
\STATE i=i+1;\\
\ENDIF
\IF{$i > 50$}
\STATE $\tau = \tau + 10^{1 + \lceil -\frac{n}{200} \rceil}$;\\
\ENDIF
\ENDFOR
\ENDFOR
\end{algorithmic}  
\end{algorithm}

However, in our implementations, $\tau=0.95$ does not suit the experiments with CIFAR100. When $\tau \in \{0.6, 0.65, 0.7, 0.75, 0.8, 0.9, 0.95\}$, PMLP performs best with $\tau \in \{0.6, 0.65, 0.7\}$. Fig.~\ref{fig520} shows PMLP's accuracy, the ratio of pseudo-labels exceeding $\tau$, and the ratio of high-confidence pseudo-labels correct with the ground truth under different fixed $\tau$ values.

The high-confidence pseudo-labels will affect the model's training by misleading the unsupervised loss $L_{U}$:
\begin{equation*}
L_{U} = I(y_{j}^{U} \geq \tau)H(Y^{U},Y^{P}),
\end{equation*}
where we encourage the prediction vector $Y^{P}$ and pseudo-label $Y^{U}$ to be consistent. 
From Fig.~\ref{fig520}, when $\tau$ is high, few high-confidence predictions are accepted and help decrease $L_{U}$. Contrastingly, a fixed low threshold $\tau$ increases the number of high-confidence pseudo-labels, but nearly half are incorrect, which introduces harmful, wrong information. 
In a word, incorrect pseudo-labels due to a fixed threshold $\tau$ negatively impact PMLP's training.

To improve PMLP's performance, we introduce an adaptive threshold policy for $\tau$, as shown in algorithm~\ref{appC}. In each training iteration, we define $r$:
\begin{equation*}
r = \frac{1}{|Y^{P}|}\sum_{j=1}^{|Y^{P}|}I(y_{j}^{U} \geq \tau),
\end{equation*}
which reflects the ratio of high-quality predictions.
PMLP generates pseudo-labels with the \textit{cluster assumption}, and its consistency is encouraged by a consistency loss $L_C$; a model is expected to create more high-quality predictions during its training. 
Thus, we can slightly increase the threshold $\tau$ to foster a more confident model. 
%In the right part of Fig.~\ref{Fig.521}, we also compare the computational costs of PMLP and LASSL to demonstrate the efficiency of PMLP. 
Fig.~\ref{fig520} also shows the superior performance of the adaptively tuning strategy. All comparative experiments are conducted on the same GPU. CIFAR10 experiments are performed on a GeForce RTX 3090, and CIFAR100 experiments are performed on a GeForce RTX 4090. 
The results show that PMLP does not significantly increase training time.

\end{document}